\documentclass{article}
\usepackage{iclr2027_conference,times}

\usepackage{amsmath,amsfonts,bm}

\def\eqref#1{equation~\ref{#1}}

\def\1{\bm{1}}

\DeclareMathAlphabet{\mathsfit}{\encodingdefault}{\sfdefault}{m}{sl}
\SetMathAlphabet{\mathsfit}{bold}{\encodingdefault}{\sfdefault}{bx}{n}

\DeclareMathOperator*{\argmax}{arg\,max}

\newcommand{\jcui}[1]{{{\textcolor{blue}{\textbf{}}}{\textcolor{blue}{\textbf{#1}}}}}
\newcommand{\jimmy}[1]{\textcolor{green}{\textbf{#1}}}

\usepackage{amsmath,amssymb,mathtools}
\usepackage{graphicx}
\usepackage{booktabs}
\usepackage{array}
\usepackage{microtype}

\usepackage[hypertexnames=false]{hyperref}
\hypersetup{
  colorlinks=true,
  linkcolor=[rgb]{0.10,0.20,0.55},
  citecolor=[rgb]{0.10,0.35,0.20},
  urlcolor=[rgb]{0.10,0.20,0.55}
}
\usepackage{url}
\usepackage{float}
\usepackage[above]{placeins}
\usepackage{subcaption}

\usepackage{xcolor}
\usepackage{tikz}
\usetikzlibrary{positioning}
\usetikzlibrary{arrows.meta}
\usepackage{multirow}
\usepackage{algorithm}
\usepackage{algpseudocode}
\usepackage{xspace}
\usepackage[most]{tcolorbox}
\usepackage{fontawesome5}

\definecolor{leapnavy}{rgb}{0.10,0.20,0.55}
\renewenvironment{abstract}{%
  \vskip.075in\centerline{\large\sc Abstract}\vspace{0.5ex}%
  \begin{tcolorbox}[enhanced,
    colback=leapnavy!3, colframe=leapnavy!75, boxrule=0.6pt, arc=2mm,
    left=4mm, right=4mm, top=2mm, bottom=2mm,
    before skip=1mm, after skip=4mm]%
}{\end{tcolorbox}}

\newcommand{\method}{LEAP\xspace}

\title{ \method: \underline{L}earned Block-wise \underline{E}vidence Retrieval for Long \underline{A}udio-Video \underline{P}erception}

\author{Juyi Lin$^{1,2}$\thanks{Corresponding author: \texttt{lin.juy@northeastern.edu}. Work done during internship at Futurewei Technologies.}, Zhiqiang Lao$^{2}$, Jiali Cui$^{2}$, Lin Zhao$^{1}$, Pu Zhao$^{1}$, Dichang Zhang$^{2}$, \\
\bf Arman Akbari$^{1}$, Yu Qi$^{1}$, Xinru Jiang$^{1}$, Yanzhi Wang$^{1}$, Heather Yu$^{2}$, Liang Peng$^{2}$ \\
$^{1}$Northeastern University \quad $^{2}$Futurewei Technologies
}

\iclrfinalcopy
\begin{document}

\maketitle
\lhead{Preprint.}%
\jcui{}%
\jimmy{}%

\vspace{-0.3in}
\begin{center}
\tcbox[on line, colback=leapnavy!6, colframe=leapnavy!75, boxrule=0.5pt, arc=2mm,
  left=2.5mm, right=2.5mm, top=1mm, bottom=1mm]{%
  \textcolor{leapnavy}{\faGlobe}\enspace\textbf{Project Page:}\enspace
  \href{https://lukelin-web.github.io/leap-website/}{\texttt{lukelin-web.github.io/leap-website}}}
\end{center}

\begin{abstract}
Hour-scale audio-visual question answering is constrained by a context dilemma: dense whole-recording encoding rapidly exhausts context limits, whereas uniform temporal compression severely dilutes fine-grained acoustic and visual evidence. We introduce \method, a framework where the model retrieves its own evidence without placing the whole recording in one context. \method divides a recording into fixed-duration blocks, applying a lightweight localization pass to each block to score short candidate windows. The highest-ranked windows are pooled and re-encoded in a single bounded answer pass. Consequently, the answer input and peak context remain independent of the recording duration. By decoupling evidence localization from reasoning, our framework can localize candidate temporal windows over pre-computed transcripts without decoding media frames, while preserving fine-grained visual and non-speech evidence by routing the final answering pass over raw audio-visual streams. \method trains both stages: a localization LoRA improves the selected windows, and an answer LoRA improves the answers read from the same windows. The block grid natively supports causal queries, enabling \method to support streaming inference without streaming-specific training. Across several AVQA benchmarks, \method improves over the Qwen3-Omni-30B-A3B baseline by 4.5--16.8\%, and transfers to a second omni-modal backbone, MiniCPM-o 4.5, surpassing its published results by 3.1--13.0\%. 
\end{abstract}

\section{Introduction}
\label{sec:intro}

Long-form audio-visual question answering (AVQA) is challenging because the evidence is often scattered across a number of brief windows within a long recording. Processing the entire recording not only consumes substantial context and memory, but also fills the model's input with largely irrelevant content.
Prior work addresses this along three axes: Selection methods keep a limited set of potentially relevant temporal regions before answering
~\citep{diao2025temporal, wang2026active, shen2025draftattention,shao2026listening}. Compression methods keep the full recording but reduce its tokens or memory states~\citep{tao2026omnizip,kong2025,zhan2024exploring,sun2026omnimem, xin2026stage}. Agentic methods revisit the source iteratively for each question~\citep{wang2026active, tao2025omniagent, zhang2025dvd}. However, selection made from a
coarse view can drop the windows an answer needs, compression thins the detail of what it keeps, and agentic methods leave where to look to general-purpose models prompted at inference, none trained to localize evidence. We therefore ask: \emph{Without full-context ingestion, can a model still pinpoint the few relevant minutes to answer a question?}

A fixed context makes coverage and density competing uses of the same tokens, which motivates visiting the timeline twice. Reading one timeline at two resolutions is an established remedy: a coarse pass decides where the evidence lies and a dense pass re-reads only those places. Prior systems scan the whole clip at low fidelity and zoom into the intervals they localize~\citep{shen2025zoom, li2026salmonnr3}, or descend recursively from long segments to short ones~\citep{hannan2025revisionllm}. When the coarse pass reads the whole recording in one context~\citep{li2026salmonnr3}, its fidelity thins with duration and it stops at the model's position limit. The scan hands the dense pass a single interval~\citep{li2026salmonnr3}, so evidence scattered over several places is out of its reach.

We introduce \method, an evidence-retrieval framework for long-form AVQA where \textbf{the answering model retrieves evidence without placing the full recording in one context}, maintaining $\mathcal{O}(1)$ context and working memory. The temporal hierarchy is fixed by duration. The learning determines which windows are kept and how to answer from them. Specifically, the recording is partitioned into non-overlapping fixed-duration \emph{blocks}, and each block is further partitioned into short \emph{candidate windows}, which serve as the basic units of evidence selection.
A lightweight, question-conditioned localization pass scores the windows within each block, and these scores are also used to rank and retain the most relevant
blocks.
\method selects only the highest-scoring windows and concatenates this small, bounded set into a single bounded \emph{answer pass}.
Each selected window is re-encoded 
at a fixed per-window context, dense enough to preserve the local details needed for answering.

Across several AVQA benchmarks, \method improves accuracy over the Qwen3-Omni-30B-A3B baseline by 4.5--16.8\%. Localization training improves the windows the retrieval stage selects, and answer training improves the answer read from those windows. \method also transfers to MiniCPM-o 4.5 \cite{cui2026minicpm}, surpassing its published results by 3.1--13.0\%.

\begin{figure}[t]
\centering
\includegraphics[width=0.9343\textwidth]{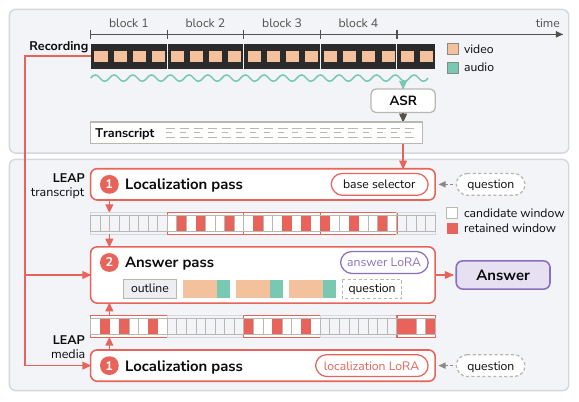}
\caption{
\textbf{Overview of \method.} \emph{Top:} each recording is split into fixed-length blocks and transcribed once. \emph{Bottom:} per question, a localization pass scores the candidate windows of every block (faint cells), from either the transcript with the base selector or the media with the localization LoRA; one answer pass re-reads the windows retained by block ranking (solid cells) from the recording.
}
\label{fig:architecture}
\end{figure}

We summarize our contributions as follows.

\begin{itemize}
\setlength{\itemsep}{2pt}
\item \textbf{Evidence retrieval without a whole-recording read.} We introduce \method, which narrows the full-length recordings 
to a bounded set of evidence-rich windows at per-pass context and working memory $\mathcal{O}(1)$ in duration.

\item \textbf{The transcript as a second scanning channel.}
Decoupling evidence localization from reasoning lets the same block grid be searched on pre-computed transcripts without decoding media frames; only the windows it selects are re-read as audio and video, so the two channels share the block grid and the answer pass and differ in what is scanned. Routing the answer pass over the raw audio-visual stream preserves the fine-grained visual and non-speech evidence a transcript misses.

\item \textbf{Historical retrieval as the stream arrives.} The block grid natively supports causal queries, enabling \method to support streaming inference without streaming-specific training (\S\ref{sec:streaming}). Localization pass selects windows over the media blocks, or over an online transcript index, up to query time, and the answer pass re-reads them from raw media.

\end{itemize}

\section{Related Work}
\label{sec:related}

\paragraph{Long-form audio-video question answering.}
Omni-modal models such as Qwen3-Omni~\citep{xu2025qwen3} jointly reason over text, video and audio. However, their finite context windows make hour-scale reasoning difficult.
Existing approaches address this problem in three main ways.
\textit{Selection-based methods} retain frames, clips, or intervals before reasoning~\citep{diao2025temporal, shao2026listening, pan2025timesearch, shen2025zoom, zhang2026timelens, hannan2025revisionllm, li2026salmonnr3}. \textit{Compression-based methods} preserve coverage while reducing visual tokens, representations, or memory states~\citep{gong2025echoingpixels, tao2026omnizip, ding2026omnisift, add-zhao-etal-2024-pruning,zhao-etal-2024-pruning,xin2026stage, kong2025, shen2024numerical,sun2026omnimem, li2024videochat, shu2025videoxl}. \textit{Agentic methods} revisit the source iteratively according to the question~\citep{tao2025omniagent, xing2026omniagent,shen2024lazydit, yang2026survey,wang2026active, zhang2025dvd, zhu2026omnirag}. 
Streaming understanding adds a causal constraint: only the recording prefix is readable when a question is asked.
Streaming systems are built to ingest each frame once and answer from what they keep, either a KV cache of the stream~\citep{di2025rekv, chen2026streamkv}, a fixed-size memory~\citep{flashvstream, zeng2026streamforest}, or a textual memory of distant history~\citep{jiang2026dynamic}. ShallowStream instead uses a shallow-layer KV cache as an index and re-processes the frames it ranks through all layers~\citep{hao2026shallowstream}.
These three families share one tension: within a fixed context, covering the recording and resolving fine evidence compete for the same tokens.
Selection can therefore drop the windows an answer needs, compression thins the detail in the windows it keeps, and agentic revisiting eases the tension by re-running its media work for every question.

\paragraph{Text as a retrieval channel.}
Text provides an efficient way to search long recordings without repeatedly processing the original media.
Caption-based pipelines aggregate descriptions of short clips to answer long-range video questions~\citep{zhang2024llovi}, while document-retrieval systems convert video into searchable text that is ultimately consumed by the answering model~\citep{ma2025drvideo}.
Other systems search a text index built over the recording while keeping the media indexed beside it, so a query that starts in captions or transcripts can drop back to frames~\citep{yin2026haven, zhang2025dvd, shen2025efficient,shen2024search,shen2025fastcar,zhan2024fast,wei2026gcr}. Others provide retrieved ASR and OCR text to the model alongside the video~\citep{luo2026videorag}.
Transcripts also serve as retrieved evidence for an omni-modal agent \citep{zhu2026omnirag}, and a long-audio planner searches timestamped streams derived from the audio, transcript among them ~\citep{someki2026planrag}.
Subtitle-based keyframe selection performs comparably to visual-search~\citep{he2026vsi}, and query-conditioned gating can select the retrieval channel or depth for each question~\citep{wang2026qgate, xue2025adavideorag}.
The caption-based, document-retrieval and long-audio planning pipelines hand their answering model text alone, never the recording itself~\citep{zhang2024llovi, shen2025sparse,zhan-etal-2024-rethinking-token,ma2025drvideo, someki2026planrag}.

\section{Methodology}
\label{sec:method}

\subsection{Problem Formulation and Overview}
\label{sec:method_overview}

Given a temporally aligned video stream $\mathcal{V}$, audio stream $\mathcal{A}$, a question $q$ and its answer options $O$, \method generate an answer $y$ without placing the complete hour-scale recording in the model's context window. As shown in Fig.~\ref{fig:architecture}, \method has two stages: one \emph{localization pass} per fixed-duration audio-video block, scoring that block's candidate windows (\S\ref{sec:coarse_glance}), and one bounded \emph{answer pass} re-encoding the high score windows (\S\ref{sec:region_readout}). Algorithm~\ref{alg:pipeline} lists the full inference procedure.

The \emph{block grid} is fixed by the clock. The stream is partitioned into non-overlapping \emph{blocks} $x_1,\dots,x_M$ of duration $\Delta=600$ s, $M = \lceil T/\Delta \rceil$ of them for a recording of length $T$. Each block is tiled into non-overlapping \emph{candidate windows} of duration $\delta=75$ s, so a full block carries $K=\lceil\Delta/\delta\rceil=8$ of them and the final block, if shorter, only its $K_m\le K$ windows that hold content. The ranking retains $B=3$ blocks, and inside each retained block the shortlist keeps $W=3$ windows, so at most $BW=9$ windows enter the answer pass. Appendices~\ref{app:selection} and~\ref{app:windows} ablate the block count, the window count and the window width. The two modalities are treated asymmetrically in the localization pass: video is compressed to a small set of visual tokens by sparse frame sampling, while all $600$ s of audio are encoded at the backbone's native rate (Appendix~\ref{app:details}). The audio keeps its full temporal support, and the visual sequence stays small enough for one inexpensive pass per block.

\subsection{Framework}
\label{sec:framework}

\method runs on two omni-modal models, Qwen3-Omni-30B-A3B~\citep{xu2025qwen3} and MiniCPM-o 4.5~\citep{cui2026minicpm}, whose weights $\theta_0$ stay frozen. An audio encoder turns each second of the waveform into $13$ tokens. A vision encoder turns the sampled frames into visual tokens. The language model reads the two token streams interleaved with the text of the prompt and produces text. A block $x_m$ encoded this way at the localization-pass media rate is the $z_m$ the localization pass reads. The trained parameters are the two LoRA adapters $\phi_{\mathrm{loc}}$ and $\phi_{\mathrm{ans}}$, each a rank-$16$ update to the query, key, value and output projections of every self-attention layer of the language model. This section's configuration values and training objective are Qwen3-Omni's; Table~\ref{tab:reprodep} lists MiniCPM-o's.

\subsubsection{Stage I: The Localization Pass and Window Scoring}
\label{sec:coarse_glance}

The frozen backbone with the localization LoRA adapter~\citep{hu2022lora} processes each block exactly once. The block's $K_m$ candidate windows are listed as lettered options in a fixed prompt; the localization pass reads the next-token logit of each option letter, and both scores use these logits,
\begin{equation}
    \left(\ell_{m,1},\ldots,\ell_{m,K_m}\right)
    = \mathcal{F}_{\theta_0,\phi_{\mathrm{loc}}}
      \!\left(z_m, q\right),
    \qquad
    r_{m,k} = \sigma\!\left(\ell_{m,k}\right),
    \qquad
    g_m = \max_{k\in\{1,\ldots,K_m\}} r_{m,k},
\label{eq:gate_score}
\end{equation}
where $z_m$ is the compressed representation of block $x_m$ and $\sigma(\cdot)$ the logistic sigmoid. A block is ranked by a monotone rescaling of its strongest candidate-window logit, a \emph{ranking} score read on the logit scale the passes share. The aggregation is a $\max$ (\S\ref{sec:loc_adapter}), ablated against a mean in Appendix~\ref{app:ranking}. The localization pass emits option-letter logits only, with no decoded answer and no cross-block key--value cache, so the peak context and memory of one localization pass depends on a single block.

\subsubsection{Stage II: Block Ranking and Answer Pass}
\label{sec:topb_gate}

After all blocks have been scanned, we retain the $\min(B,M)$ highest-scoring blocks under $g_m$, restored to chronological order before answer generation. The whole pipeline costs $N_{\mathrm{fwd}} = M + 1$ passes per question, with no dependence on $B$ (Appendix~\ref{app:efficiency}).

\label{sec:region_readout}
Within each retained block the shortlist keeps its $\min(W,K_m)$ highest-scoring windows under $r_{m,k}$. Every block's shortlist is formed during its own localization pass, from the window scores Eq.~\ref{eq:gate_score} has already produced. The retained blocks' shortlists are pooled into the evidence of the answer pass.

Only these selected windows are reloaded from the original recording. Sorted by absolute time into windows $(J_1,\ldots,J_R)$ with $R\leq BW$, they form the bounded answer-pass sequence
\begin{equation}
    \bigl[
        \omega;\,
        c_0;\,
        \tau_1;\,
        \operatorname{Enc}(J_1);\,
        \cdots;\,
        \tau_R;\,
        \operatorname{Enc}(J_R);\,
        q;\,
        O
      \bigr],
\label{eq:readout_input}
\end{equation}
where $\omega$ is the transcript outline introduced below, $c_0$ is a fixed instruction stating that the segments are discontinuous and must be reasoned over jointly, $\tau_i$ carries window $i$'s absolute timestamp, and $\operatorname{Enc}(J_i)$ encodes the synchronized video and audio of the $i$th window's temporal support $J_i$. Each retained window is re-read from the source at the \emph{per-window context}: $32$ video frames per $75$\,s window plus that window's audio at the backbone's native rate.

\paragraph{The transcript outline.} The answer pass also reads $\omega$, a text outline of the whole recording, built from a single transcription made once, before any question is asked: one line per minute of speech, each carrying its minute mark, capped at $4{,}000$ tokens. A transcription longer than the cap is filled question-first, the minutes matching the question and its options admitted ahead of the rest (Appendix~\ref{app:details}). A recording without speech contributes no outline.

The backbone swaps the localization adapter $\phi_{\mathrm{loc}}$ for the answer adapter $\phi_{\mathrm{ans}}$; each is trained from the frozen base backbone on its own data, and the two are never stacked. The answer pass generates over all selected windows jointly, so evidence from different blocks interacts in one bounded context.

\subsection{Localization Adapter}
\label{sec:loc_adapter}

The localization adapter is trained on this task, on questions derived from LongVALE~\citep{geng2025longvale}, whose event annotations supply the evidence span. Every training clip fits in one block. For a clip with annotated evidence span $J^{\star}$, the supervised letter is the candidate that best covers it. Supervision is the cross-entropy over the same option-letter logits the localization pass reads,
\begin{equation}
\mathcal{L}_{\mathrm{loc}}
= -\!\!\!\sum_{\left(z,q,J^{\star}\right)\in\mathcal{D}_{\mathrm{loc}}}\!\!\!
  \log\frac{\exp\!\left(\ell_{k^{\star}}\right)}
            {\sum_{k=1}^{n}\exp\!\left(\ell_{k}\right)},
  \qquad
  k^{\star} = \argmax_{k}\,\left|J_k \cap J^{\star}\right|.
\label{eq:localization_loss}
\end{equation}
Here $z$ is a training clip encoded at the localization-pass media rate, $q$ its localization question, and $J^{\star}$ its annotated evidence span. The sum runs over the $n$ lettered candidates $J_1,\ldots,J_n$ of one supervision level (Table~\ref{tab:repro}). These candidates are set by the training clip rather than by $\delta$: every training clip is shorter than one block, and tiling it at $\delta$ would leave most clips with fewer than the eight letters a deployed localization pass reads. A longer clip is therefore first cut into eight equal candidate windows, carrying whole-clip audio and sparse video as a deployed localization pass does, so its first level reads the same eight letters; a second, training-only level covers the finer windows inside the one that holds the evidence. A shorter clip is tiled directly into at most eight finer windows.

Training applies the cross-entropy objective of Eq.~\ref{eq:localization_loss} to the single target $k^\star = \arg\max_k \vert{}J_k \cap J^\star\vert{}$, the window that overlaps the evidence most. MiniCPM-o 4.5's selector is instead trained with overlap-fraction BCE, a binary cross-entropy that grades every window by its overlap with the evidence (Table~\ref{tab:reprodep}); on Qwen3-Omni that objective ranks blocks worse across passes, yet does not yield a statistically detectable difference in final accuracy (Appendix~\ref{app:ranking}). Inference scores a block by its maximal candidate-window posterior, $g_m = \max_k \sigma(\ell_k)$. Replacing the $\max$ by a mean discards the margin by which the block's best window stands above the other windows of its pass, the component the block ranking runs on, and costs most of the evidence coverage (Appendix~\ref{app:ranking}).

\subsection{Answer Adapter}
\label{sec:ans_adapter}

The answer adapter is trained via cross-entropy over the answer token and the end-of-turn token, conditioned on an evidence sequence $\mathcal{X}$ of the same form but without the outline $\omega$,
\begin{equation}
    \mathcal{L}_{\mathrm{ans}}
    = -\frac{1}{|y|}\sum_{t=1}^{|y|}
      \log p\!\left(y_t\mid y_{<t},\mathcal{X}\right).
\label{eq:answer_loss}
\end{equation}
The sequence of Eq.~\ref{eq:readout_input} is the inference-time instance of $\mathcal{X}$, with $\omega$ added. Each training instance stitches the gold evidence segment with distractor segments drawn from \emph{other} recordings, rendered in the same form as the retrieved sequence: timestamped raw audio-video segments, one of which carries the evidence. Full setup is in Appendix~\ref{app:repro}.

\subsection{Advantages}

\paragraph{Bounded context and peak memory.}
Either pass reads a context of fixed size: the localization pass reads one block, and the answer pass at most $BW$ selected windows together with the outline and the question, so peak context and memory per pass are $\mathcal{O}(1)$ in $T$. One pass over the whole recording instead costs $\approx$$47$k tokens per hour: the backbone's $65{,}536$-token position limit is exhausted at roughly $81$ minutes. Appendix~\ref{app:efficiency} details the token accounting.

\paragraph{Localization training within one block.}
The objective of Eq.~\ref{eq:localization_loss} is defined within one block, so training, like inference, never reads more than one block, whereas selectors trained by reinforcement from the answer~\citep{pan2025timesearch,li2026salmonnr3} carry the whole recording and a live answering model in every update. Its supervision marks where the evidence lies, never answer correctness, so the selector never learns which inputs one answering model gets right (Appendix~\ref{app:priorsearch}).

\paragraph{Coverage and detail on separate passes.}
\method separates covering the recording from resolving its evidence: the localization pass covers every block, and the answer pass spends its context on the few retained windows, at a per-window context that does not change with $T$. Whichever channel selects those windows, media or transcript, the answer pass re-encodes the same raw audio-video windows and reads the transcript only as a short outline. What a transcription misses, unspoken visual detail or non-speech audio, therefore stays readable at answering time.

\section{Experiments}
\label{sec:results}

\subsection{Settings}

We evaluate \method on four main benchmarks: TraceAV-Bench~\citep{feng2026traceav}, LVOmniBench~\citep{tao2026lvomnibench}, VideoOdyssey-AV~\citep{he2026videoodyssey} (videos all longer than 60 minutes), and MMOU~\citep{goel2026mmou}. We also evaluate OmniVideoBench~\citep{li2025omnivideobench}, included in the paired ablations of this section, and on three video-only benchmarks, LVBench~\citep{wang2025lvbench}, CG-Bench mini~\citep{chen2025cg} and Video-MME~\citep{fu2025video}, reported in Appendix~\ref{app:length}. Fine-grained video evaluation has also become increasingly important beyond semantic understanding, extending to structured assessment of physical reasoning in generative world models~\citep{lin2026phygroundbenchmarkingphysicalreasoning,rupprecht2026human}.
The block configuration is unchanged across all datasets and carried as is to StreamArena~\citep{streamarena2026} in \S\ref{sec:streaming}. On the four main benchmarks, CG-Bench and StreamArena the answer pass over the selected windows carries the transcript outline of \S\ref{sec:outline}; the selector comparisons of Figure~\ref{fig:locablation} and Table~\ref{tab:retrieverkey} answer without it. The \emph{base selector} is the same selection run by the \emph{base model}, the backbone with no adapter mounted. An effect is \emph{significant} when its paired video-clustered interval excludes zero (Appendix~\ref{app:protocol}). Appendix~\ref{app:contamination} reports the contamination audit.

\begin{table}[t]
\centering
\caption{\textbf{Main results on Qwen3-Omni-30B.} Accuracy (\%). Upper block: closed-source systems (\textsuperscript{$\circ$}) as published. Middle block: open systems as published, then our backbone; \textsuperscript{*} = the published number for our backbone; \emph{caption cascade} = audio and visual captions read by a text-only model; \textsuperscript{\ddag} = the benchmark's official input recipe re-run on our stack, or an official-style whole-clip run where none is runnable. Lower block: \emph{\method without retrieval} = \method's answer LoRA reading the whole clip in one pass (on VideoOdyssey, its audio-montage recipe); \emph{\method without answer training} = \method's localization LoRA also answering the selected windows; --- = no number. TraceAV = the mean over its twelve general sub-tasks. \textbf{Bold} = best, \underline{underline} = second best in each column.}
\label{tab:main}
\small
\setlength{\tabcolsep}{4pt}
\begin{tabular}{lcccc}
\toprule
System & TraceAV & LVOmni & VideoOdyssey (1--4h) & MMOU \\
\midrule
gemini-3-flash-preview\textsuperscript{$\circ$} & \textbf{62.3} & \textbf{59.0} & 44.3 & --- \\
Qwen3.5-Omni-Plus\textsuperscript{$\circ$} & --- & --- & 43.0 & --- \\
\midrule
Ming-Flash-Omni-2.0 & 51.7 & 34.6 & --- & --- \\
Caption cascade into Qwen3-235B & --- & --- & --- & 47.9 \\
Qwen3-Omni-30B, as published\textsuperscript{*} & 48.4 & 35.8 & 28.7 & 54.1 \\
Qwen3-Omni-30B, with ASR transcript\textsuperscript{*} & --- & 42.2 & --- & --- \\
Qwen3-Omni-30B, official recipe & 56.4\textsuperscript{\ddag} & 40.7\textsuperscript{\ddag} & 36.9\textsuperscript{\ddag} & 56.5\textsuperscript{\ddag} \\
~~Whole clip, uniform, frame-matched & 53.6 & 40.4 & --- & 57.8 \\
\midrule
\method without retrieval & 54.8 & 42.9 & 41.9 & \underline{61.8} \\
\method without answer training & 60.1 & 43.0 & \underline{46.7} & 61.4 \\
\method & \underline{60.9} & \underline{45.8} & \textbf{53.7} & \textbf{66.3} \\
\bottomrule
\end{tabular}
\end{table}

\subsection{Main Results}

\begin{figure}[t]
\centering
\includegraphics[width=0.9\textwidth]{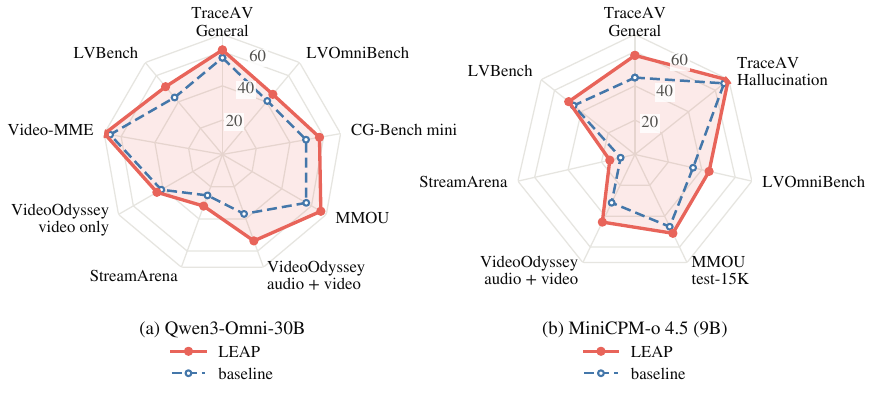}
\caption{\textbf{Across benchmarks and backbones.} Solid = \method, dashed = baseline: \emph{(a)} the same backbone on the whole recording (the causal prefix on StreamArena); \emph{(b)} the backbone as published (official-style on our stack for VideoOdyssey and LVBench, native streaming on StreamArena).}
\label{fig:benchradar}
\end{figure}

Table~\ref{tab:main} compares \method against same-stack baselines using either the official input recipe or an official-style whole-clip run (Appendix~\ref{app:samestack}). Our method leads significantly on all four. \method also leads the whole clip read, sampled uniformly and frame-matched baselines. Figure~\ref{fig:benchradar}(a) extends the comparison beyond the four main benchmarks, reading LVBench and Video-MME with audio removed as their official protocols do, and our method leads on all of them.

\subsection{Retrieval Stage}
\label{sec:selection}

\begin{figure}[t]
\centering
\includegraphics[width=0.9\textwidth]{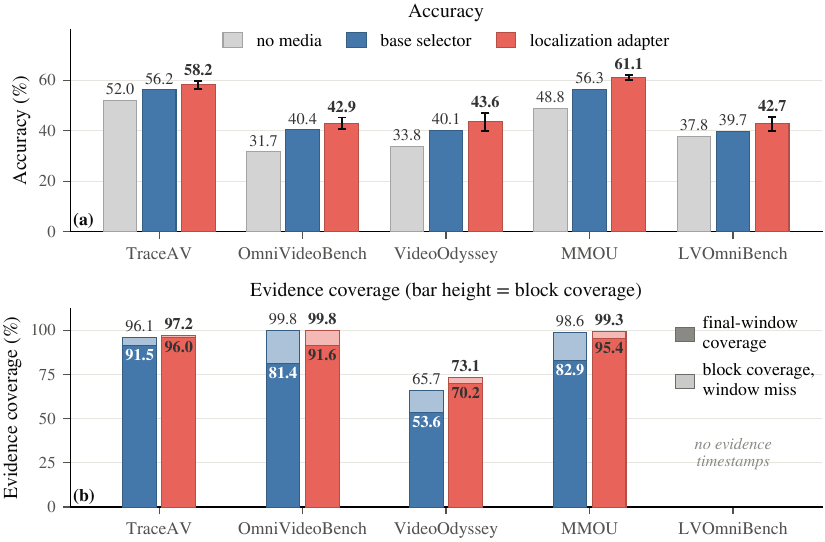}
\caption{\textbf{The retrieval stage ablated.} \emph{no media}: stem and options only. \emph{(a)} Accuracy; the localization adapter's whisker (paired, vs.\ base selector) clears the base-selector bar where significant. \emph{(b)} Evidence coverage: full bar = in a retained block, saturated = in an answer-pass window; OmniVideoBench on its timestamped subset, MMOU where both selectors stored a selection.}
\label{fig:locablation}
\end{figure}

Figure~\ref{fig:locablation} shows localization training raises \emph{evidence coverage}, the share of questions whose annotated evidence the selection retains: substantially in the answer-pass windows, and in the retained blocks most on VideoOdyssey. How much of that added coverage turns into accuracy differs across benchmarks: \method has almost no evidence coverage left to gain on TraceAV, and by far the most on VideoOdyssey. The trained selector also beats an untrained ranking score: an off-the-shelf retriever or the confidence of an answer read on each block put in its place, everything else held, covers less of the evidence and answers less accurately, pooled over the benchmarks of Table~\ref{tab:retrieverkey}.

\method gains from where it places its windows: it answers significantly more accurately on TraceAV and VideoOdyssey than when the same windows are redrawn at random positions over the recording (Appendix~\ref{app:randomwindow}). It also gains from its block ranking: on VideoOdyssey it is significantly ahead of equally spaced blocks in both accuracy (Figure~\ref{fig:selection}) and evidence coverage (Appendix~\ref{app:ranking}).

\subsection{Answer Stage}
\label{sec:answerstage}

Table~\ref{tab:formatcontrol} fixes the answer LoRA and changes only what it reads: the whole clip, or \method's windows. \method leads on all four benchmarks. Part of that lead comes from clips too long to encode whole. Without those questions, \method keeps most of its lead on LVOmniBench, still significant. Against the frame-matched whole clip, which reads as many frames as \method but is answered by the base model, \method leads significantly on TraceAV, LVOmniBench and MMOU (Appendix~\ref{app:uniform}). \method transfers to MiniCPM-o 4.5 with the same block grid and backbone-specific adapter training (Table~\ref{tab:reprodep}). \method raises evidence coverage and accuracy significantly on VideoOdyssey and MMOU (Appendix~\ref{app:answer}).

\subsection{Audio and Transcripts}
\label{sec:audiolevels}

\method reads audio in both passes. On TraceAV, silencing only the \emph{localization pass} changes the selected windows substantially. Muting only the answer pass costs accuracy significantly on the hearing-required and cross-modal question classes (Table~\ref{tab:audioablation}).

\paragraph{Transcript channel.}
\label{sec:audiofirst}
The media channel spends one localization pass on the video and audio of every block, speech often provides sufficient discriminative information. We therefore transcribe each recording once, before any question, and build a second channel that selects windows from the transcript alone. Each candidate window is represented by the transcript of its span and scored by the base selector, with the media channel's prompt and the same block-then-window selection.

\begin{table}[t]
\centering
\caption{\textbf{Transcript-based retrieval against the media channel.} Accuracy (\%). Both channels feed the deployed answer pass. \emph{Localization tokens} count the localization passes only; \emph{seconds} = question-to-answer wall clock on one idle GPU over a duration-stratified sample, transcription charged to neither. TraceAV = accuracy over all its questions, hallucination sub-tasks included.}
\label{tab:zoomend}
\small
\begin{tabular}{@{}lcccc@{}}
\toprule
System & TraceAV & LVOmni & VideoOdyssey (1--4h) & MMOU \\
\midrule
Media channel & 63.6 & 45.8 & 53.7 & 66.3 \\
Transcript channel & 64.6 & 46.5 & 52.3 & 66.2 \\
\midrule
\multicolumn{5}{@{}l}{\emph{Localization tokens per question, media channel $\div$ transcript channel}} \\
Media $\div$ transcript & 5.41$\times$ & 4.74$\times$ & 6.21$\times$ & 4.93$\times$ \\
\midrule
\multicolumn{5}{@{}l}{\emph{Query-time seconds per question (median)}} \\
Media channel & 12.4 & 15.6 & 30.0 & 5.5 \\
Transcript channel & 7.6 & 11.0 & 18.6 & 3.4 \\
\bottomrule
\end{tabular}
\end{table}

Both channels hand their windows to the deployed answer pass (Table~\ref{tab:zoomend}). The transcript channel spends a fraction of the media channel's localization tokens and answers faster on every benchmark. The more of the annotated evidence is spoken, the more of it the transcript channel retains relative to the media channel (Appendix~\ref{app:textregime}). \method deploys the media channel throughout; where a recording's evidence is spoken, the transcript channel is the cheaper substitute.

\paragraph{Transcript outline.}
\label{sec:outline}
\method's answer pass also reads a minute-stamped outline of the same transcript. Table~\ref{tab:outline} replaces the deployed outline by one that thins every minute equally, or removes it. Removing the outline costs \method accuracy significantly on TraceAV, VideoOdyssey and MMOU. On the questions whose outline is cut, \method's question-first outline beats one thinning every minute equally, significantly on VideoOdyssey. \method's gain from the outline is largest per question where the retained windows miss the evidence (Appendix~\ref{app:outline}).

\begin{figure}[t]
\centering
\includegraphics[width=0.9\textwidth]{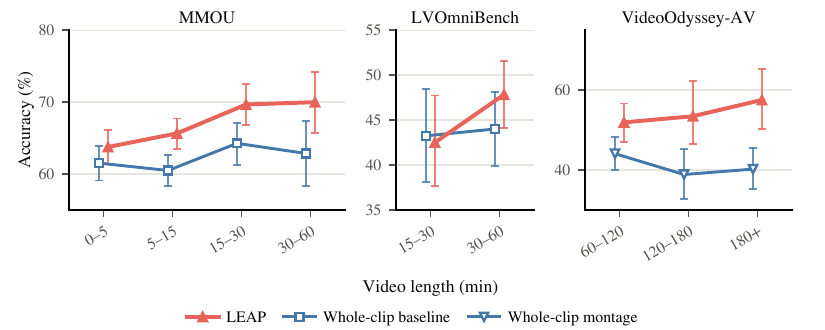}
\caption{\textbf{Accuracy vs.\ video length.} All panels share one grid of length buckets, and each shows the buckets it populates. Error bars are each line's video-clustered 95\% interval.}
\label{fig:length}
\end{figure}

\subsection{Duration and Evidence Distance}

\method's accuracy shows no systematic collapse as videos get longer (Figure~\ref{fig:length}), and on MMOU it rises with duration. \method and a single whole-recording pass spend their context differently as the duration $T$ grows. A baseline that fits the whole recording into one fixed context must spread that context over all of it, so what it reads per minute of source falls as $1/T$. \method reads every retained window at the same per-window context at any duration.
\label{sec:beyond}
A strong baseline is the context-filled montage (Appendix~\ref{app:protocol}), which packs the whole recording into one bounded pass. \method leads significantly in every quintile, including the one where the montage keeps almost all the audio (Figure~\ref{fig:montage_cov}).

\begin{table}[t]
\centering
\caption{\textbf{Historical retrospection under causal access on StreamArena, Qwen3-Omni-30B.} Accuracy (\%). Bins = minutes from the query time back to the evidence. \emph{Whole prefix} reads everything before the query time in one pass; \emph{Uniform windows} fills the same context with windows spread equally over it; \emph{\method{} -- media} and \emph{\method{} -- transcript} select the windows from the media blocks or from the transcript index built as the stream arrives, and answer them from the raw media. \textbf{Bold} = best, \underline{underline} = second best in each column.}
\label{tab:streamarenamain}
\small
\begin{tabular}{@{}lrrrrr@{}}
\toprule
& & \multicolumn{4}{c}{evidence distance (min)} \\
\cmidrule(lr){3-6}
System & HR avg & $\le$5 & 5--15 & 15--30 & $>$30 \\
\midrule
Whole prefix & 25.4 & 22.1 & 34.2 & 25.5 & 12.6 \\
Uniform windows & 20.2 & 23.2 & 25.8 & 17.9 & 9.3 \\
\method{} -- media & \textbf{32.0} & \textbf{26.5} & \textbf{41.2} & \underline{29.3} & \underline{23.6} \\
\method{} -- transcript & \underline{30.1} & \underline{24.3} & \underline{36.1} & \textbf{30.4} & \textbf{24.7} \\
\bottomrule
\end{tabular}
\end{table}

\paragraph{Causal access.}
\label{sec:streaming}
StreamArena~\citep{streamarena2026} asks each question at a moment of a stream, lets the model read only what came before that moment, and bins the questions by how far back the evidence lies. \method's block grid runs on it unchanged, with no streaming training. The answer pass reads the retained windows with the base weights and the transcript outline, and always includes the window ending at the query time (Table~\ref{tab:reprodep}). The block grid is fixed by the clock and every candidate window is scored inside its own block (Appendix~\ref{app:streamarena}). \method significantly leads the same backbone reading the whole prefix in one pass and windows spread equally over the prefix (Table~\ref{tab:streamarenamain}).

The transcript channel of \S\ref{sec:audiofirst} suits a live stream. The stream is transcribed in time order with no lookahead, a question selects windows from the transcript available at its query time, and the answer pass re-reads those windows from the raw media. Its windows contain the annotated evidence about as often as those of the media channel. Answering them from their transcript instead of the raw media costs accuracy significantly in all but the farthest distance bin. On the 9B MiniCPM-o 4.5, \method also beats the model's native streaming mode significantly, overall and in the second distance bin.

\section{Conclusion}
\label{sec:conclusion}

Long-form audio-visual question answering is limited by a fixed context: spread over hours, it leaves too little detail per minute. We introduced \method, an evidence-retrieval framework that reads the recording in bounded pieces: every pass reads one fixed-duration block or a bounded set of retained windows, so peak context and memory are independent of duration. Across four benchmarks \method improves over each benchmark's official-style recipe on the same backbone by 4.5--16.8\%, shows no systematic collapse with duration, and transfers to a second omni-modal backbone. The same block grid searches a timestamped transcript with 4.7--6.2$\times$ fewer localization tokens than the media channel, and applies to a causal prefix without streaming training. Open directions include stronger retrieval scores within the bounded-cost structure, and an answer pass that can tell when its retained windows miss the evidence and return to the scan for more.

\section*{Reproducibility statement}

\method is specified in \S\ref{sec:method} and step by step in Algorithm~\ref{alg:pipeline}. Both adapters, their training pools and the inference settings are listed in Table~\ref{tab:repro}, and those of the second backbone in Table~\ref{tab:reprodep}; Appendix~\ref{app:details} gives how the localization pool is built, the optimizer, the prompts, the letter read-out and the transcript rendering. Appendix~\ref{app:protocol} fixes the evaluation protocol of our runs, including the denominators and the clustered confidence intervals; Appendices~\ref{app:samestack} and \ref{app:streamarena} describe how the same-stack baselines and the causal-access protocol are run, and Appendix~\ref{app:contamination} audits the training sources against the benchmarks. All benchmarks and training corpora are public releases. Code will be released upon acceptance of the paper.

\bibliography{references}
\bibliographystyle{iclr2027_conference}

\appendix
\FloatBarrier
\section{The Method in Detail}
\label{app:methoddetail}

\subsection{The inference algorithm}
\label{app:arch}

See Algorithm \ref{alg:pipeline}.

\begin{algorithm}[htbp]
\caption{\method inference.}
\label{alg:pipeline}
\begin{algorithmic}[1]
\Require video $V$ with its audio (any duration), question $q$, options $O$
\State $\omega \gets \textsc{Outline}(V, q, O)$ \Comment{one transcription per recording, its minutes ranked for the question}
\State $x_1,\dots,x_M \gets \textsc{TileNonOverlapping}(V, 600\text{s})$
\For{$m = 1,\dots,M$} \Comment{exactly one localization pass per block}
  \State $\ell \gets \textsc{LocalizationPass}(x_m, q)$ \Comment{up to 8 window letters}
  \State $\mathcal{U}_m \gets \textsc{Top}(\ell, W{=}3)$;\ \ $g_m \gets \max_{r \in \mathcal{U}_m} \sigma(\ell_r)$
\EndFor
\State $\mathcal{S} \gets \textsc{Top}(\{g_m\}, B{=}3)$ \Comment{selected blocks}
\State $\mathcal{U} \gets \textsc{TimeSortedUnion}(\{\mathcal{U}_m : m \in \mathcal{S}\})$ \Comment{$\le BW{=}9$ windows by construction}
\State \Return $\textsc{Answer}(\textsc{PackWindows}(\mathcal{U}, \omega, q, O))$ \Comment{per-window context, one bounded pass}
\end{algorithmic}
\end{algorithm}

\subsection{Training and inference configuration}
\label{app:repro}

\begin{table}[htbp]
\centering
\caption{\textbf{Configuration of the two adapters.}}
\label{tab:repro}
\footnotesize
\begin{tabular}{@{}lp{\dimexpr\textwidth-114pt-2\tabcolsep\relax}@{}}
\toprule
\multicolumn{2}{@{}l}{\emph{Backbone and adapters}} \\
\midrule
Backbone & Qwen3-Omni-30B-A3B-Instruct, frozen in both stages \\
LoRA rank / scaling & $r=16$ / $\alpha=32$, identical for both adapters \\
Attachment points & attention query, key, value and output projections \\
Composition & sequential, one adapter resident at a time, never stacked \\
Training hardware & H100, bf16 throughout \\
\midrule
\multicolumn{2}{@{}l}{\emph{Localization adapter $\phi_{\mathrm{loc}}$}} \\
\midrule
Objective & cross-entropy over the option-letter logits against the best-covering candidate, Eq.~\ref{eq:localization_loss}, at each supervision level listed below \\
Data & LongVALE-derived localization questions, filtered by the base model (Appendix~\ref{app:details}) \\
Training clip length & $30$--$598$\,s; one $600$\,s block each \\
Supervision grid & clips up to $240$\,s: one level of $30$\,s windows; longer clips: two levels, first $8$ equal windows spanning the clip, then the $30$\,s windows inside the one that holds the evidence \\
Learning rate & $1\times10^{-4}$, cosine without warmup \\
Items per optimizer step & $8$ \\
Deployed checkpoint & step $2000$ \\
\midrule
\multicolumn{2}{@{}l}{\emph{Answer adapter $\phi_{\mathrm{ans}}$}} \\
\midrule
Data & $3{,}404$ instances, gold segments from OmniVideo-100K~\citep{cai2026omnivideo}, over $1{,}022$ distinct gold source videos, plus $175$ held-out instances \\
Instance construction & $3$ segments per instance (gold $+2$ distractors drawn from \emph{other} recordings); the source question's $4$ options \\
Target and loss masking & bare gold option letter plus the end-of-turn token; loss on those tokens only, per-item target-token mean \\
Learning rate & $1\times10^{-4}$, cosine without warmup to zero at step $900$ \\
Items per optimizer step & $12$ \\
Deployed checkpoint & step $900$ \\
\midrule
\multicolumn{2}{@{}l}{\emph{Deployed inference}} \\
\midrule
Block length & $\Delta=600$\,s, non-overlapping \\
Localization-pass media & video at $0.5$ fps capped at $24$ frames; all block audio at native rate \\
Candidate windows & $\delta=75$\,s each \\
Selection & top $B=3$ blocks $\times$ top $W=3$ windows, so $\leq BW=9$ windows \\
Answer-pass media & $32$ frames per window (up to $2$ fps) plus that window's audio at native rate \\
Answer-pass transcript outline & question-ranked minute lines, at most $4{,}000$ tokens \\
Answer-pass position cap & $36{,}864$ tokens \\
Decoding & greedy, at most $256$ new tokens \\
\bottomrule
\end{tabular}
\end{table}

\begin{table}[htbp]
\centering
\caption{\textbf{MiniCPM-o 4.5 and the causal-access protocol.}}
\label{tab:reprodep}
\footnotesize
\begin{tabular}{@{}lp{\dimexpr\textwidth-114pt-2\tabcolsep\relax}@{}}
\toprule
\multicolumn{2}{@{}l}{\emph{MiniCPM-o 4.5}} \\
\midrule
Backbone & MiniCPM-o 4.5 \\
Attachment points & attention query, key, value and output projections, and the MLP gate, up and down projections \\
Localization objective & overlap-fraction BCE over up to eight $75$\,s candidate windows \\
Localization supervision grid & one level for every clip \\
Localization items per step & $12$ \\
Data-parallel ranks & $6$ for the localization adapter, $3$ for the answer adapter \\
Answer data & $13{,}192$ instances, gold segments from OmniVideo-100K and LongVALE, over $4{,}215$ distinct gold source videos \\
Answer deployed checkpoint & step $1100$, where the cosine schedule ends \\
Localization-pass media & video at $1$ fps capped at $320$ frames per block \\
Answer-pass media & $10$ frames per window \\
Answer-pass transcript outline & none \\
\midrule
\multicolumn{2}{@{}l}{\emph{Causal-access protocol on StreamArena}} \\
\midrule
Input & the causal prefix of the recording up to the query time \\
Answer weights & the base model, no adapter mounted \\
Answer-pass transcript outline & built from the text transcribed by the query time, question-ranked, at most $4{,}000$ tokens \\
Reserved window & the $75$\,s ending at the query time, added to the retained windows (replacing the lowest-scoring one when the context is full) unless a retained window already overlaps it at an intersection-over-union of at least one half \\
Decoding & free text; the generation itself is the answer, with no letter parser \\
\bottomrule
\end{tabular}
\end{table}

Table~\ref{tab:repro} lists both adapters' configuration and the deployed inference settings; Table~\ref{tab:reprodep} lists where MiniCPM-o 4.5 and the causal-access protocol of Appendix~\ref{app:streamarena} depart from them. On MiniCPM-o 4.5 the answer pass reads fewer frames per second than the localization pass, ten per window being where this backbone's accuracy stops rising on long recordings. Re-answered with the localization adapter mounted over the same retained windows, more frames per window, up to one per second, do not raise accuracy on LVOmniBench or OmniVideoBench, and one per second lowers it.

\subsection{Data, prompts, and renderers}
\label{app:details}

\paragraph{Localization corpus.} Every LongVALE event segment yields one question: the stem asks in which time window the segment's caption occurs, the segment's span is the annotated evidence span $J^{\star}$, and the clip is its source video, which never exceeds one block. The candidate windows are enumerated at training time from the evidence span and the clip duration on the grids of Table~\ref{tab:repro}, and the target at each level is the candidate overlapping the evidence span most, the earlier one on ties. An evidence span that starts inside the clip's first $3$\,s or first $5\%$ is dropped. The quality filter is the base model: shown one candidate window at a time, it is asked whether the evidence for the event appears in that segment, and a question is kept only when it says yes on the target candidate and no on every other candidate; an unparseable reply counts as no. The corpus keeps $40{,}284$ of $69{,}630$ questions, over $4{,}735$ source videos. Videos are split first, by a hash of their identifier, one in ten to validation.

\paragraph{Optimization.} Both adapters are trained by one trainer: AdamW with $\beta=(0.9,0.95)$, $\epsilon=10^{-8}$ and decoupled weight decay $0.01$, learning rate $10^{-4}$ under a cosine schedule without warmup, gradient norm clipped at $1$, bf16 with gradient checkpointing, two data-parallel ranks, the gradient averaged over the items of a step. The localization adapter's two supervised levels each contribute one cross-entropy of Eq.~\ref{eq:localization_loss}, summed with equal weight. Localization checkpoints are scored every $500$ steps on a fixed $247$-question set from the validation split by whether the evidence span's midpoint falls inside one of the three $30$\,s windows the localization pass shortlists on the two-level grid of Table~\ref{tab:repro}, the best $30$\,s window inside each of its three highest-scoring first-level windows; step $2{,}000$ scores best and is deployed. The answer adapter runs $900$ steps, about three epochs over the instances.

\paragraph{Prompts and the letter read-out.} The localization pass is one user turn with no system message: the block's video, the block's audio, then the text below, where the windows are listed in absolute clip time and the question is the benchmark's own stem; the question's answer options are not shown.
\begin{quote}\small\ttfamily
In which time window does the evidence to answer this question appear?\\
\{question\}\\
A. 10:00\textrm{--}11:15\\
B. 11:15\textrm{--}12:30\\
\ldots\\
H. 18:45\textrm{--}20:00\\
Answer with the option's letter from the given choices directly.
\end{quote}
The hidden state at the last prompt position is read against the output-head rows of the lettered options alone, each letter tokenized in its prompt context, and those logits are the $\ell_{m,k}$ of Eq.~\ref{eq:gate_score}.

The answer pass is also one user turn: the transcript outline where the recording has speech, then a line saying several discontinuous segments follow, each labelled with its clip timestamp, to be reasoned over jointly, then the retained windows in time order, each introduced by \texttt{Segment $i$ [$s_0$s-$s_1$s]:} carried as a video clip with its audio, then the question. On TraceAV the question closes with the benchmark's official prompt, which asks for the letter alone and for comma-joined letters when several options are correct. The reply is decoded greedily and parsed by its first standalone letter among the question's option letters; on TraceAV the final letter group is read as a set and must match the key exactly. A reply with no option letter is scored wrong.

\paragraph{Media context.} Both passes cap every frame at $336^2$ pixels, so a 16:9 source decodes at $448{\times}252$, which the backbone's processor raises to its minimum frame area at $512{\times}288$: $72$ visual tokens per frame ($144$ per two-frame pair), the figure Appendix~\ref{app:efficiency} uses; audio costs $13$ tokens per second at either stage. The localization pass samples $0.5$\,fps up to $24$ frames, so a block of $48$\,s or more carries $24$ frames and a shorter one $0.5$\,fps of it, never fewer than $4$; a full block costs about $9{,}700$ tokens, of which $7{,}800$ are audio. The answer pass samples $2$\,fps up to $32$ frames per window, so a window of $16$\,s or more carries $32$ frames and a shorter one $2$\,fps of it, never fewer than $4$; frame counts are even throughout. Audio is always the whole span at the native rate. A recording tiles into $\lceil T/600\,\mathrm{s}\rceil$ blocks and a block into windows, the last of each keeping the remainder, so a trailing window is at most $75$\,s and a trailing block of under $75$\,s holds one candidate.

\paragraph{The transcript channel's rendering.} A candidate window is rendered as the transcript segments overlapping it, in time order, cut at $1{,}200$ characters: segments are appended whole until the next would overflow, and the rest of the window is replaced by an ellipsis; a window with no speech reads ``(no speech)''. The block is rendered as one line per lettered window carrying its time range, followed by the window-selection prompt the media channel uses, with no media element in the conversation, and scored by the base model from the same option-letter logits.

\paragraph{The transcript outline.} The transcript is produced once per recording by Whisper large-v3 under \texttt{faster-whisper}, greedy with voice-activity filtering and the language detected automatically, as segments carrying start and end times. Segments are bucketed into minutes by their start time and joined into one line per minute headed by that minute's range; a minute with no speech produces no line. To rank the minutes against a question, the stem and every option text are lower-cased and split into alphanumeric words. A minute scores the sum, over the query words it contains, of $\log(N/\mathrm{df})$, with $N$ the number of spoken minutes of that recording and $\mathrm{df}$ the number of them containing the word, so a word spoken in every minute weighs nothing. An outline that fits the $4{,}000$-token cap is passed verbatim. Otherwise every minute header is kept, and the minutes with a positive score are admitted whole in descending score (the earlier minute first on ties) while each fits the space that remains. The space left after that is shared by the minutes not admitted: each keeps the same leading fraction of its own tokens, closed by an ellipsis, so no minute disappears. Tokens are counted by the backbone's tokenizer, lines are emitted in time order under a one-line outline header.

\subsection{Cost accounting: a bounded peak footprint}
\label{app:efficiency}

Let $n_q$ be the number of question and instruction tokens, $n_{\omega}$ the transcript outline's token cap, and $n_{\delta}$ the maximum token count of one selected window at the per-window context. Since at most $BW$ windows enter the answer stage,
\begin{equation}
    n_{\mathrm{ans}}
    \leq n_q + n_{\omega} + BW n_{\delta}+n_{\mathrm{sep}},
\label{eq:context_bound}
\end{equation}
where $n_{\mathrm{ans}}$ is the answer-pass length in tokens and $n_{\mathrm{sep}}$ accounts for timestamps and separators. Every term on the right is a configured constant, with no term in $T$, and the localization-pass length is independently bounded by one compressed block.

For Qwen3-Omni, the terms of Eq.~\ref{eq:context_bound} translate to tokens as follows: one answer-pass frame costs $72$ tokens at the deployed pixel cap (Appendix~\ref{app:details}) and audio $\approx$$13$ tokens per second, so one selected $75$\,s window carries $n_{\delta}\approx3{,}300$ tokens and the nine-window answer pass $\approx$$30{,}000$, inside the deployed answer-pass cap of $36{,}864$ tokens, itself well under the backbone's $65{,}536$-token position limit. The transcript outline adds at most $4{,}000$ tokens, counted in every answer-pass figure below.

\emph{Query work} is one localization pass per block plus the single bounded answer pass, so the pass count is fixed at $M{+}1$ (\S\ref{sec:topb_gate}): median $13$ passes on VideoOdyssey and single-digit medians elsewhere. In the deployed runs, every block pass stays under $10{,}000$ tokens on Qwen3-Omni and under $30{,}000$ on MiniCPM-o 4.5, and no answer pass reaches the deployed position cap. \emph{Cumulative} prefill tokens are $2.2$--$2.4\times$ those of a single whole-clip pass over the same questions (Qwen3-Omni medians on TraceAV, LVOmniBench and a stratified MMOU subset). \emph{Working memory} is the peak of one pass: the weights plus the key--value state of at most one answer pass at the position cap, so \textbf{peak context and peak memory are bounded at any duration}.

\paragraph{The accounting is per-question worst case: the media prefix is reusable.} The block tiling and each block's fixed grid of candidate windows are set by the clip alone, the backbone is frozen, and every localization-pass conversation places the block's media tokens \emph{before} the question text. The media decode, the encoder features, and the media-prefix KV of every block pass are therefore identical by construction across questions about the same video, and the question suffix is ${\approx}2\%$ of a localization pass. These prefixes are precisely the artifacts the prefix-reuse literature caches and serves~\citep{promptcache, radixattention, cachegen}, and that streaming video models keep across questions as the stream's KV cache~\citep{di2025rekv, chen2026streamkv}.

Paying for each block's media prefix once per \emph{video}, with every block prefix of the video kept until its last question, takes the median per-question cumulative prefill on VideoOdyssey ($10.6$ questions per video) from $148.1$k to $42.5$k tokens ($0.29\times$). This reuse trades memory for prefill: the cache holds one localization-pass prefix per block, so it grows linearly with duration and lies outside the per-pass memory bound above, which the deployed runs meet with no cache. Such memory–throughput tradeoffs are particularly relevant in multimodal deployment, where model architecture and hardware platform can substantially affect both peak memory footprint and inference throughput~\citep{lin2026votevisionlanguageactionoptimizationtrajectory, taherin2026cross, NEURIPS2025_baca5eb9,ng2024open}.
Only the localization pass reuses; the answer pass reads per-question windows. A vLLM serving run shows the reuse is realizable. We run the localization pass of six $1.5$--$2.1$\,h videos through a four-way tensor-parallel vLLM server whose key--value pool holds one video's block prefixes. Automatic prefix caching hits the cached block prefixes at $87.9\%$ and cuts the median localization-pass wall-clock of a video's later questions by $1.6\times$ against the same server with caching disabled.

Question-in-the-loop re-observation~\citep{pan2025timesearch, wang2026active} decides \emph{what media to process} from the question, so its media work re-runs per question by construction. Memory-bank streaming models~\citep{moviechat, malmm} also ingest once, into a lossy fixed-size state; \method's cached block prefixes are exact encodings whose size grows with the recording, and the selected windows are re-read at the per-window context.
\subsection{Relation to prior search over long recordings}
\label{app:priorsearch}

First, training matches inference in being $\mathcal{O}(1)$ context in duration: Equation~\ref{eq:localization_loss} is defined within one block, so no training step reads past one block. Selection policies optimized by reinforcement from the answer~\citep{pan2025timesearch,li2026salmonnr3} instead carry a live answering model through every update. Frame-Voyager keeps that model offline, ranking frame combinations by its loss once and training the selector on the frozen labels, but its selector still reads candidate frames drawn from across the whole recording in one context~\citep{yu2025frame}. Second, the supervision never sees answer correctness, whereas a selector trained on answer reward learns which inputs make one particular answering model produce a response that is marked correct.

Cross-block ranking needs no training of its own: every pass shares one prompt and one adapter, and the blocks are ranked on those logits directly (Appendix~\ref{app:ranking}). The scorer is the answering backbone itself: relevance is an option-letter logit read from the frozen backbone under the localization adapter. No separate retriever, captioner or embedding index stands beside the model, and ranking reuses logits the localization pass has already produced: ranking adds no parameters, no pass, and no term in $B$ on top of the $M{+}1$ cost. The score is question-conditioned and jointly audio-visual with full-rate audio, where previous selectors score one modality~\citep{wang2026active,shao2026listening}, reach the audio through the visual track~\citep{diao2025temporal}, or return a single interval~\citep{zhang2026timelens}.

Finally, no pass over a recording longer than one block holds the whole recording in context: two-stage systems process the full recording in their first stage~\citep{hannan2025revisionllm,li2026salmonnr3} and agentic methods re-observe the source per question~\citep{tao2025omniagent,xing2026omniagent,zhang2025dvd}; in our method every pass reads one block or a bounded set of retained windows.

\method differs from Zoom-Zero's divide-and-conquer variant~\citep{shen2025zoom}, which also splits a recording into non-overlapping windows scanned independently, in two ways. First, our window score is an option-letter logit the localization pass already produced, so ranking adds no pass; its score is a trial answer's confidence, which costs a second pass per block and varies mostly between questions, the part of a score a within-question ranking cannot use (Appendix~\ref{app:ranking}). Second, our scan is jointly audio-visual, reading each block's audio at the native rate, while its scan is visual.

\FloatBarrier
\section{Evaluation Protocol and Baselines}
\label{app:eval}

\subsection{Systems, contexts, and reporting conventions}
\label{app:protocol}

\paragraph{The baselines.} Table~\ref{tab:comparators} lists the single-pass baselines on Qwen3-Omni-30B and what each reads. The \emph{context-filled montage} stands in for the whole clip wherever one pass has to fit a recording of any length: $32$ frames spread over the recording plus uniform $10$\,s audio slices, packed until the encoded input, media and question together, reaches $60{,}000$ tokens, leaving room for the reply under the $65{,}536$-token position limit. A short recording therefore keeps nearly all of its audio and a four-hour one keeps a fraction. On VideoOdyssey the official recipe is our approximation of the benchmark's $64$-frame and audio-montage pipeline (Appendix~\ref{app:samestack}).

\begin{table}[htbp]
\centering
\caption{\textbf{The single-pass baselines on Qwen3-Omni-30B.} Each reads the whole recording, or on StreamArena the causal prefix, in one answer pass; \emph{uniform windows} instead keeps the answer pass of \method and places its windows at equal spacing. \emph{Frames} are spread evenly over what the baseline reads; \emph{full track} = the native waveform of that span; \emph{official montage} = $64$ ten-second audio slices at evenly spaced starts, concatenated into one track, as cut by the benchmark's released evaluation code. The official recipes of TraceAV and LVOmniBench also carry the benchmark's own prompt.}
\label{tab:comparators}
\footnotesize
\renewcommand{\arraystretch}{1.3}
\begin{tabular}{@{}>{\raggedright\arraybackslash}p{3.5cm}>{\raggedright\arraybackslash}p{2.3cm}>{\raggedright\arraybackslash}p{2.4cm}>{\raggedright\arraybackslash}p{\dimexpr\textwidth-8.2cm-6\tabcolsep\relax}@{}}
\toprule
Baseline & Frames & Audio & Read in \\
\midrule
\multicolumn{4}{@{}l}{\emph{Official recipe}\textsuperscript{\ddag}} \\
~~TraceAV & up to $256$ & full track & Table~\ref{tab:main}, Figure~\ref{fig:benchradar}a \\
~~LVOmniBench & $128$ at $336^2$ & full track & Table~\ref{tab:main}, Figure~\ref{fig:benchradar}a \\
~~VideoOdyssey & $64$ & official montage & Tables~\ref{tab:main} and~\ref{tab:vo_tasks}, Figure~\ref{fig:benchradar}a \\
~~MMOU & $32$ & full track & Table~\ref{tab:main}, Figure~\ref{fig:benchradar}a \\
\midrule
Frame-matched whole clip & $288$ & full track & Tables~\ref{tab:main} and~\ref{tab:concurrent}, Appendix~\ref{app:uniform} \\
Whole clip & $32$ & full track (none on LVBench and Video-MME in Figure~\ref{fig:benchradar}a) & Tables~\ref{tab:main} and~\ref{tab:formatcontrol}, Figures~\ref{fig:benchradar}a, \ref{fig:length}, \ref{fig:length_appendix} and~\ref{fig:niah} \\
Context-filled montage & $32$ & $10$\,s slices up to $60{,}000$ tokens & Table~\ref{tab:concurrent}, Figures~\ref{fig:length}, \ref{fig:length_appendix} and~\ref{fig:montage_cov} \\
Native-sampling whole clip & $2$\,fps, at most $128$ & official montage, or none & Table~\ref{tab:vo_ccl}, Figures~\ref{fig:benchradar}a and~\ref{fig:vo_cw} \\
Whole prefix & $32$ & full track & Table~\ref{tab:streamarena}, Figure~\ref{fig:benchradar}a \\
Uniform windows & $32$ per window, up to $9$ windows & each window's own & Table~\ref{tab:streamarena} \\
\bottomrule
\end{tabular}
\end{table}

\paragraph{The whole-clip limit.} Many of these videos are too long for a single whole-clip pass: some TraceAV and LVOmniBench questions, and the large majority of VideoOdyssey questions, exceed the whole-clip limit, the longest recording that fits one whole-clip pass inside the $65{,}536$-token position limit, roughly $81$ minutes at the media context used here ($32$ frames plus full audio). OmniVideoBench and MMOU are the only benchmarks of \S\ref{sec:results} entirely within it. \method's answer pass runs under a tighter cap of $36{,}864$ tokens (Table~\ref{tab:repro}), which its nine windows never reach (Appendix~\ref{app:efficiency}).

\paragraph{Reporting conventions.} Percentage-point
differences are written \emph{pp}. Every paired difference in this paper is read on one inferential quantity:
the two-sided $95\%$ bootstrap interval on the paired difference,
resampled at the source video, since many questions share one recording, at a
fixed seed and at least $10{,}000$ resamples. An effect is
\emph{significant} exactly when that interval excludes zero. The error
bars on the length curves of Figures~\ref{fig:length} and~\ref{fig:length_appendix} are per-line intervals resampled
at the source video in the same way, and carry no paired claim. Unless a caption says otherwise, every denominator is the full official
question set, with failures scored as errors where gold labels are public.

\subsection{Concurrent systems on these benchmarks}
\label{app:concurrent}

\begin{table}[t]
\centering
\caption{\textbf{Concurrent systems on the same backbone.} Accuracy (\%). The first two rows read the same input: the whole recording sampled uniformly and frame-matched to the most frames our answer pass reads, with its full audio, or on VideoOdyssey the context-filled audio montage. \emph{OmniVideo-100K SFT} = the released checkpoint fine-tuned on that instruction set, no retrieval stage \citep{cai2026omnivideo}. \emph{OmniRAG-Agent loop} = the unmodified backbone inside that paper's released retrieval agent, scored by our answer extractor \citep{zhu2026omnirag}. \textsuperscript{*}~= the benchmark's own runs, as TraceAV reports them \citep{feng2026traceav}: the bare backbone, and two untrained agents driving it, \emph{AVP} \citep{wang2026active} and \emph{Audio-Guided} OmniAgent \citep{tao2025omniagent}. --- = no number. TraceAV columns follow its two published tables: subtask macro-averages over the general and over the hallucination sub-tasks. \textbf{Bold} = best, \underline{underline} = second best in each column.}
\label{tab:concurrent}
\small
\setlength{\tabcolsep}{4pt}
\begin{tabular}{@{}lccccc@{}}
\toprule
System & TraceAV & TraceAV & LVOmni & VideoOdyssey (1--4h) & MMOU \\
 & General & Hallucination &  &  &  \\
\midrule
Qwen3-Omni-30B & \underline{53.6} & 68.0 & \underline{40.4} & 37.9 & 57.8 \\
~~+ OmniVideo-100K SFT & 52.4 & 65.0 & 39.3 & \underline{43.1} & \textbf{70.1} \\
\midrule
~~+ OmniRAG-Agent loop & 39.5 & 63.4 & 27.5 & 26.6 & 34.8 \\
\midrule
Qwen3-Omni-30B, as published\textsuperscript{*} & 48.4 & 67.5 & --- & --- & --- \\
~~+ AVP\textsuperscript{*} & 47.1 & \underline{69.8} & --- & --- & --- \\
~~+ Audio-Guided\textsuperscript{*} & 42.4 & 57.9 & --- & --- & --- \\
\midrule
\method{} & \textbf{60.9} & \textbf{76.2} & \textbf{45.8} & \textbf{53.7} & \underline{66.3} \\
\bottomrule
\end{tabular}
\end{table}

Video-SALMONN-R\textsuperscript{3}~\citep{li2026salmonnr3} trains a two-pass re-watch policy by reinforcement learning on a Qwen3-VL backbone with a separate audio encoder, and OmniRAG-Agent~\citep{zhu2026omnirag} reports OmniVideoBench on a self-selected subset; its reinforcement-trained version exists only on smaller backbones, with no released weights, so we run its untrained version on our stack (Table~\ref{tab:concurrent}). Agentic pipelines~\citep{tao2025omniagent} run frontier-API components once per question, and offline index-and-search systems~\citep{yin2026haven} add a per-video indexing cost. OmniVideo-100K~\citep{cai2026omnivideo} also fine-tunes omni backbones on cross-segment audio-visual evidence chains, but as whole-clip instruction data with no retrieval at inference. \method's answer adapter trains on instances from its gold segments (Table~\ref{tab:repro}) and reads the windows a trained selector retrieves (\S\ref{sec:ans_adapter}).

OmniVideo-100K's released 30B checkpoint is our own backbone fine-tuned that way, so we run it on our stack under our own protocol (Table~\ref{tab:concurrent}). \method leads it on every benchmark except MMOU. On MMOU, \method's retrieval on that checkpoint lifts it significantly, from $70.1$ to $71.6$: the checkpoint answers the windows \method selects, with \method's transcript outline, instead of the whole clip.

The OmniRAG-Agent loop wraps the same unmodified backbone in an external index: the recording is time-compressed to a fixed-length clip placed in the first turn, and any evidence beyond that clip arrives as CLIP-ranked frames and speech-transcript segments fetched over a bounded sequence of tool calls. \method outperforms OmniRAG-Agent on all four benchmarks.

TraceAV's own appendix runs two further untrained agents on this backbone, AVP~\citep{wang2026active} and Audio-Guided OmniAgent~\citep{tao2025omniagent}, with the backbone standing in for the models each was released with (Table~\ref{tab:concurrent}). \method leads both agents on both of its tables.

\subsection{The same-stack official-recipe baselines}
\label{app:samestack}

Where the published same-backbone number was produced on a pipeline we cannot reproduce bit-for-bit, we instead re-run the official \emph{input recipe} on our own stack (same backbone weights, same harness) and read method gain only against that same-stack baseline. These are the \textsuperscript{\ddag} cells of Table~\ref{tab:main}. Paired against them on the same questions, \method without answer training leads significantly on TraceAV, VideoOdyssey and MMOU. On VideoOdyssey that baseline re-implements the official audio-montage API pipeline on our stack. With that montage (Table~\ref{tab:comparators}) the model hears about eleven minutes of any recording. Our re-implementation departs from the official pipeline in three ways.

\paragraph{Audio.} The model hears the complete montage waveform. The official pipeline base64-encodes a $16$\,kbps MP3 of the ${\approx}640$s montage into an OpenAI-compatible \texttt{input\_audio} field.

\paragraph{Frames.} We pass the $64$ frames as a \emph{video} element, so the model sees temporal order and adjacent-frame pairing. The official pipeline sends the same $64$ frames as $64$ independent still images side by side, with no timestamps and no video-side temporal encoding.

\paragraph{Denominator.} The official protocol moves unanswered or unparsed questions \emph{out} of the denominator, so the published figure is accuracy over the answerable subset only.

Smaller divergences remain: a forced-guess instruction in the official prompt, message ordering, decoding temperature, and JPEG/MP3 lossy round-trips.

\paragraph{The official recipes of LVOmniBench and MMOU.} LVOmniBench's official pipeline runs on HF transformers with greedy decoding and the full native waveform, with more frames than ours. It still needs a same-stack baseline, on two counts. \textbf{The official recipe does not fit the backbone it is run on}: the official model wrapper pins neither the frame count nor the per-frame resolution, so both fall to the video-loading library's defaults, sized for a $128$k-token context while this backbone's position limit is $65{,}536$. Re-encoding the official conversation verbatim puts every question past the limit: the $768$-frame video stack alone exceeds it before a single audio token is added. \textbf{And the official harness drops failures from the denominator}, filtering out every question whose pass failed (out-of-memory included) or whose answer failed to parse, with no skip count reported.

An official-style base whole-clip read on our stack (official prompt, full native waveform, greedy, no adapter, $128$ frames at $336^2$) is LVOmniBench's baseline in \S\ref{sec:results}.

MMOU publishes a scoring protocol but no input recipe, so its baseline is an official-style whole-clip read on the base model at our media context, greedy, one pass in the original option order, whereas the official scoring uses five option permutations, majority voting, and best-of-prompt selection.
\subsection{The frame-matched whole clip}
\label{app:uniform}

\begin{table}[t]
\centering
\caption{\textbf{Input-form control.} \method and the whole clip carry the same answer LoRA and no transcript outline, and differ in input form: the whole clip is read up to the backbone's position limit, the selected windows at the answer-pass cap. Under \emph{full denominator} a clip too long to encode whole inside the backbone's position limit is scored wrong for the whole clip; \emph{both forms fit} drops those questions. TraceAV = accuracy over all its questions, hallucination sub-tasks included. VideoOdyssey is absent: most of its recordings are too long to encode whole.}
\label{tab:formatcontrol}
\small
\begin{tabular}{lcccc}
\toprule
 & \multicolumn{2}{c}{Full denominator} & \multicolumn{2}{c}{Both forms fit} \\
\cmidrule(lr){2-3}\cmidrule(lr){4-5}
Benchmark & Whole clip & Selected windows & Whole clip & Selected windows \\
\midrule
TraceAV & 57.3 & 60.4 & 59.9 & 60.6 \\
LVOmni & 42.9 & 47.5 & 43.7 & 47.3 \\
OmniVideoBench & 42.3 & 43.3 & 42.3 & 43.3 \\
MMOU & 61.8 & 65.2 & 61.8 & 65.2 \\
\bottomrule
\end{tabular}
\end{table}

Table~\ref{tab:formatcontrol} pairs \method against the whole clip at our own media context of $32$ frames plus the full audio. A second control matches the \emph{frames} instead: the whole clip answered at \method's own ceiling of $288$ frames ($9$ windows $\times$ $32$), sampled evenly across the recording and carrying the full native waveform, and answered by the base model. Its audio contains everything \method's does, since \method's windows carry no more audio than the whole track. At $288$ frames some clips no longer fit: questions on TraceAV, LVOmniBench and MMOU overflow and are scored wrong.

It is read against \method without answer training. On the full denominator, where a question the frame-matched whole clip cannot fit is scored wrong for it, \method without answer training leads significantly on TraceAV and MMOU, and on MMOU also once those questions are dropped. \method itself leads significantly on TraceAV, LVOmniBench and MMOU on both denominators.

\subsection{The causal-access protocol on StreamArena}
\label{app:streamarena}

\begin{table}[t]
\centering
\caption{\textbf{Historical retrospection under causal access on StreamArena.} \emph{HR} bins = minutes from the query time back to the evidence, under the benchmark's own bin names. \emph{Query-time input} = what the model reads when the question arrives. Published rows are the benchmark's own table under its judge: \textsuperscript{$\circ$} = closed source, \textsuperscript{a} = marked there as an author-finetuned backbone. \emph{\method{} -- media} scans the media blocks with the localization adapter mounted, \emph{Base selector} the same blocks without it, and \emph{\method{} -- transcript} the transcript index built as the stream arrives; all three answer the retained windows from the raw media. \emph{Transcript only} answers the windows of \emph{\method{} -- transcript} from their transcript instead. \textbf{Bold} = best, \underline{underline} = second best in each column within a group.}
\label{tab:streamarena}
\footnotesize
\setlength{\tabcolsep}{1.5pt}
\begin{tabular}{@{}lllrrrrr@{}}
\toprule
& & & & \multicolumn{4}{c}{evidence distance (min)} \\
\cmidrule(lr){5-8}
& & & & L1 & L2 & L3 & L4 \\
System & Backbone & Query-time input & HR avg & $\le$5 & 5--15 & 15--30 & $>$30 \\
\midrule
\multicolumn{8}{@{}l}{\emph{Published, benchmark judge}} \\
StreamMind & Qwen3.5-397B-A17B & continuous memory & \underline{34.9} & \textbf{31.5} & \textbf{46.7} & \textbf{34.6} & \underline{17.1} \\
Qwen3.5-Omni\textsuperscript{$\circ$} & --- & whole prefix & \textbf{35.8} & \textbf{31.5} & \underline{44.8} & \underline{34.1} & \textbf{25.4} \\
AURA\textsuperscript{a} & Qwen3-VL-8B & recent window & 22.7 & \underline{25.4} & 27.0 & 24.3 & 10.5 \\
MiniCPM-o-4.5\textsuperscript{a} & Qwen3-8B & recent window & 9.8 & 10.5 & 13.3 & 7.6 & 5.0 \\
StreamForest & Qwen2-7B & whole prefix, compressed & 14.4 & 19.3 & 13.6 & 14.1 & 11.0 \\
\midrule
\multicolumn{8}{@{}l}{\emph{Same stack, local judge, no streaming training}} \\
Whole prefix & Qwen3-Omni-30B & the whole prefix, one pass & 25.4 & 22.1 & 34.2 & 25.5 & 12.6 \\
Uniform windows & Qwen3-Omni-30B & $\le 9$ uniform windows & 20.2 & 23.2 & 25.8 & 17.9 & 9.3 \\
\method{} -- media & Qwen3-Omni-30B & $\le 9$ windows, media scan & \textbf{32.0} & \textbf{26.5} & \textbf{41.2} & \underline{29.3} & \underline{23.6} \\
Base selector & Qwen3-Omni-30B & $\le 9$ windows, media scan & 27.1 & 23.8 & 32.1 & 27.2 & 21.4 \\
\method{} -- transcript & Qwen3-Omni-30B & $\le 9$ windows, transcript scan & \underline{30.1} & \underline{24.3} & \underline{36.1} & \textbf{30.4} & \textbf{24.7} \\
Transcript only & Qwen3-Omni-30B & same windows, as text & 22.3 & 14.4 & 27.9 & 22.8 & 19.8 \\
\midrule
Native streaming & MiniCPM-o 4.5 & last 30 s & 8.6 & 13.3 & 7.3 & 4.9 & 9.9 \\
\method & MiniCPM-o 4.5 & $\le 9$ selected windows & 15.1 & 18.2 & 18.8 & 10.3 & 9.9 \\
\bottomrule
\end{tabular}
\end{table}

\paragraph{The benchmark.} StreamArena~\citep{streamarena2026} poses open-ended questions at timestamped moments of hour-long recordings and scores a free-text answer by whether it contains the reference answer's factual core. We report its historical-retrospection task, every question answered independently from the recording alone. Every pass's input ends at the query time, so the $600$\,s block grid tiles exactly the causal prefix, a median of $4$ and at most $12$ blocks per question.

\paragraph{The same-stack comparison on Qwen3-Omni-30B.} \method and the whole-prefix baseline share the frozen backbone, the prompt (stating the query time and that \emph{now} is the last segment's end) and the judge, a local Qwen3.6-27B applying the benchmark's binary criterion with one vote.

\begin{figure}[htbp]
\centering
\includegraphics[width=\textwidth]{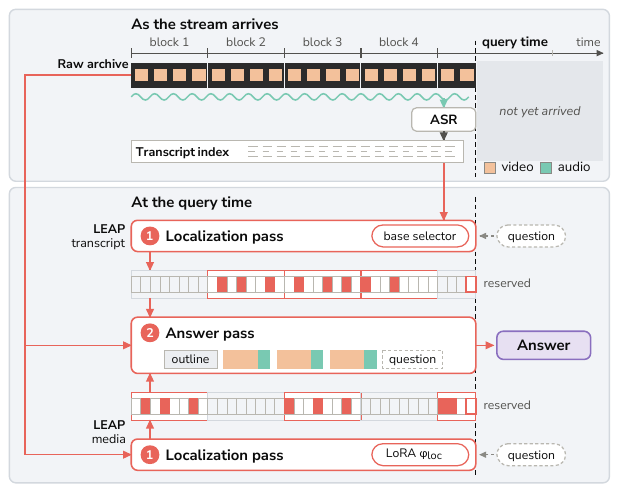}
\caption{\textbf{\method in the streaming scenario.} One time axis, cut by the query time; nothing to its right exists when the question is asked. \emph{Top}: the raw archive and the transcript index built beside it. \emph{Bottom}: the two selectors of \method and the one answer pass they share, which reads the retained windows from the archive and the outline from the index. Faint cells are candidate windows, solid cells the windows retained inside the blocks each selector keeps, and the outlined cell the slot held for the moment before the query, whether or not the ranking kept its block.}
\label{fig:stream}
\end{figure}

\method here is the deployed pipeline with no streaming training (Figure~\ref{fig:stream}): the localization adapter scans the causal blocks at the localization-pass context, retains the top $3$ and shortlists $3$ windows each, and the base model answers the retained windows at $32$ frames per $75$\,s window with the windows' own audio. The window ending at the query time, or a retained window overlapping it at an intersection-over-union of at least one half, is always among them (Table~\ref{tab:reprodep}).

The whole-prefix baseline answers the whole causal prefix in one pass (the whole-clip recipe: $32$ frames spread over the prefix with its full audio); questions whose prefix does not fit are scored as errors. Restricting the pairing to the questions whose prefix fits leaves the lead significant ($+5.60$\,pp).

The uniform-windows baseline keeps \method's context (up to $9$ windows of $75$\,s) and replaces the ranking by windows spread equally over the causal prefix, the last one ending at the query time; a prefix that fits that context is read whole. It reads at least as many windows as \method, and more on nearly every question where \method under-fills its context, and is answered by the same weights, prompt and judge. \method's ranked windows cover the annotated evidence on nearly three quarters of the historical-retrospection questions, against about half for the uniform windows.

\paragraph{A transcript index built as the stream arrives.} The index is built in time order (Figure~\ref{fig:stream}). faster-whisper large-v3 transcribes the audio in $30$\,s chunks with no lookahead and no overlap, each chunk decoded with only the previous chunk's text as context. Every transcribed segment is stamped with the moment its text became available under a real-time schedule: one ASR worker per stream, a chunk decoded once its audio is complete. A question reads only the segments available at its query time. The base selector scores every causal block's candidate windows from that text, as the transcript channel does offline, and its retained windows go to the same answer pass, the base model, with the same reserved window for the most recent $75$\,s. On one L40S the transcription runs at a real-time factor of $0.028$, and text arrives a median $0.8$\,s after the chunk's audio ends ($1.3$\,s at the $95$th percentile). The index occupies $70$\,KB per recording hour, beside $0.88$\,GB of raw archive.

Its windows cover the annotated evidence on $70.9\%$ of the historical-retrospection questions, against $72.1\%$ for the media scan of \method. With the evidence modality labeled by the judge model from each question and its annotated evidence description, the transcript channel covers spoken evidence significantly more often than that media scan and visual evidence slightly less often.

The two scans differ in the selector as well as in the index, so a further comparison holds the selector fixed: the base selector scans the media with the localization adapter unmounted, and its windows go to the same answer pass (\emph{Base selector} in Table~\ref{tab:streamarena}). Under that one selector the transcript channel covers the evidence significantly more often than the media, by $20.6$\,pp, in every distance bin and for visual as well as spoken evidence, and its answers are significantly more often correct. Without the localization adapter the windows cover about as much of the evidence as the uniform ones; block coverage stays close to the adapter's, and the gap is in the choice of window inside a retained block. Mounting the localization adapter closes that gap, raising media coverage by $21.8$\,pp.

Answering the retained windows from the raw media leads answering from their transcript alone significantly in all but the farthest distance bin, and leads the uniform windows (Table~\ref{tab:streamarena}). Heard and not seen (each retained window as its audio alone), the same windows recover $3.5$\,pp of the gain over the transcript; the frames on top of the audio add the remaining $4.2$\,pp, and both parts are significant.
\paragraph{The transcript outline at the answer pass.} \method's answer pass reads the transcript outline of \S\ref{sec:outline}, cut at the query time. Without it, \method's accuracy barely moves overall.

\paragraph{Re-reading the selected windows at the published resolution and frame rate.} The published offline rows read up to $128$ frames at up to 720p and the streaming rows one or two frames per second. Re-reading the selected windows of \method at 1280$\times$720 with $12$ frames per $75$\,s window, trading frames for resolution, or at $80$ frames per window ($1.07$\,fps) at the deployed resolution barely moves historical-retrospection accuracy against the deployed per-window context over the same windows, all three read without the outline. Both re-reads run under a position cap widened to $65{,}536$ tokens, since the denser windows overflow the deployed cap. Cutting the answer pass to the localization pass's own frames, about three per window with the audio unchanged, costs $2.5$\,pp significantly.

\begin{figure}[htbp]
\centering
\includegraphics[width=0.5\textwidth]{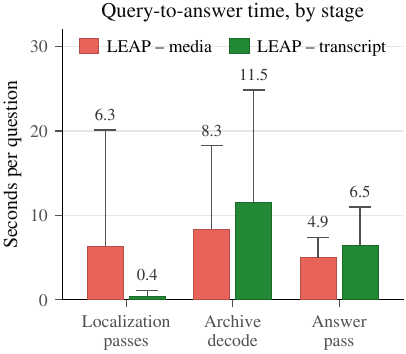}
\caption{\textbf{Query-to-answer time by stage on StreamArena.} Seconds per historical-retrospection question on one H100, one group per pipeline stage: the localization passes over the causal blocks, decoding the retained windows from the archive, and the answer pass. The bar is the per-question median and the whisker above it reaches the 95th percentile.}
\label{fig:stream_latency}
\end{figure}

\paragraph{Latency.} Figure~\ref{fig:stream_latency} splits query-to-answer time on a single H100, with no prefix caching and no batching, into every causal block's localization pass, the media decoding of the retained windows, and the one answer pass. Under the transcript index nearly all of the localization time disappears, and decoding and answering remain. Localization's tail grows with the number of causal blocks before the query, and the transcript index removes that tail along with the median.

\paragraph{The same-stack comparison on MiniCPM-o 4.5.} The smaller backbone answers this benchmark at the media context of Table~\ref{tab:reprodep} and with no streaming training, and it supplies a baseline the larger one cannot, its native streaming mode: half-duplex over the $30$\,s before the query time, one frame paired with one second of $16$\,kHz audio per second. Both conditions answer with the released weights (\method's selector carries the localization adapter of Table~\ref{tab:reprodep}), greedy, one local judge, every question answered alone, over the same questions and clustered interval as the Qwen3-Omni-30B group. \method leads the native streaming mode inside the same-stack group (Table~\ref{tab:streamarena}); the published rows are scored under the benchmark's judge with the earlier turns of its multi-turn protocol in context.

\subsection{Contamination}
\label{app:contamination}

The two source corpora of the deployed training pools (LongVALE and OmniVideo-100K, $12{,}454$ videos, a superset of every video behind either adapter's training items) were audited against every benchmark but StreamArena at three levels where the source media are available: video identifier; perceptual frame hash on a ${\approx}5$\,s grid, the $300$ pairs per benchmark with the most offset-consistent matches re-scored by grayscale normalized cross-correlation; and Haitsma--Kalker audio fingerprints matched under offset consistency. On TraceAV, whose recordings mostly carry a YouTube identifier, the frame and audio levels cover the recordings without one.

At the identifier level, no benchmark video meets a training set of the backbone it is reported for.

\FloatBarrier
\section{The Retrieval Stage}
\label{app:retrieval}

\begin{table}[t]
\centering
\caption{\textbf{Removing \method's components.} Accuracy (\%) on Qwen3-Omni-30B. Each row removes one component \textbf{from the row above}, and the answer LoRA answers in every row. $-$ \emph{transcript outline} answers the same windows without the outline; $-$ \emph{localization LoRA} lets the bare backbone's coarse scan pick the windows; $-$ \emph{retrieval} reads the whole clip in one pass (on VideoOdyssey, its audio-montage recipe). TraceAV = the mean over its twelve general sub-tasks. \textbf{Bold} = best, \underline{underline} = second best in each column.}
\label{tab:componentablation}
\small
\begin{tabular}{@{}lcccc@{}}
\toprule
System & TraceAV & LVOmni & VideoOdyssey & MMOU \\
\midrule
\method & \textbf{60.9} & \underline{45.8} & \textbf{53.7} & \textbf{66.3} \\
~~$-$ transcript outline & \underline{57.8} & \textbf{47.5} & \underline{51.2} & \underline{65.2} \\
~~~~$-$ localization LoRA & 56.2 & 43.5 & 46.7 & 62.1 \\
~~~~~~$-$ retrieval & 54.8 & 42.9 & 41.9 & 61.8 \\
\bottomrule
\end{tabular}
\end{table}

Table~\ref{tab:componentablation} removes \method's components one at a time. Reading the whole clip in place of the retrieved windows lowers accuracy significantly on TraceAV and VideoOdyssey. The windows \method's localization LoRA selects are answered significantly better than those of the base selector on LVOmniBench, VideoOdyssey and MMOU. \method gains significantly from the transcript outline on TraceAV, VideoOdyssey and MMOU.

\subsection{Selection interventions: the retained-block count}
\label{app:selection}

\begin{figure}[t]
\centering
\includegraphics[width=0.86\textwidth]{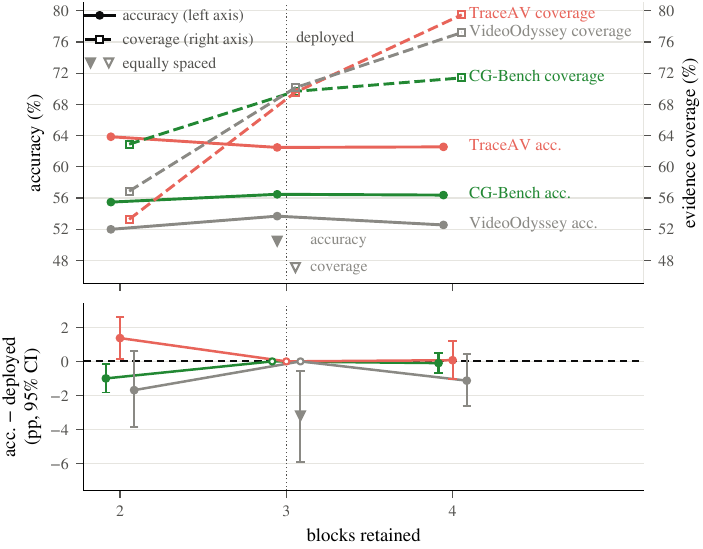}
\caption{\textbf{Accuracy and evidence coverage against the retained-block count.} \emph{Top}: levels, accuracy on the left axis and evidence coverage on the right; triangles hold the deployed count and replace the block ranking by equally spaced blocks. \emph{Bottom}: the paired accuracy difference against the deployed count, with its bootstrap interval; the deployed count is drawn open at zero. TraceAV runs only on the recordings that tile into more blocks than the deployed count keeps.}
\label{fig:selection}
\end{figure}

Unless a paragraph says otherwise, every ablation in this appendix re-ranks or re-answers the stored per-window logits of the deployed localization passes. A paragraph that runs a new localization pass (another selector, a silenced or single-channel input, a retrained adapter) says so where it appears. The trial-answer variance of \S\ref{app:ranking} reads a separate TraceAV localization pass, the \emph{$30$\,s-window scan}: the same $600$\,s blocks with $30$\,s candidate windows, under a prompt that also offered a `no evidence' option. The CG-Bench panel of Figure~\ref{fig:selection} and the CG-Bench window-count ablation of \S\ref{app:windows} read a CG-Bench localization pass under that same prompt with the `no evidence' option. Table~\ref{tab:retrieverkey} reads localization passes that split a recording's partial tail block differently: a tail block of up to $240$\,s is one candidate window, a longer one eight equal windows. The two selector variants of Figure~\ref{fig:locablation} share the localization pass, window enumeration and block ranking, so they differ only in the localization adapter. The answer passes of Figures~\ref{fig:locablation} and~\ref{fig:loc_conversion}, of Table~\ref{tab:retrieverkey}, of the CG-Bench panel of Figure~\ref{fig:selection} and of the window-count and window-width ablations of \S\ref{app:windows} read no transcript outline.

\begin{table}[t]
\centering
\caption{\textbf{An off-the-shelf retriever as the ranking score.} Every row keeps the same candidate-window grid, block and window counts and answers with the localization adapter mounted; rows differ only in the score a candidate is ranked by. Under the trial-answer score the windows inside a block are still ranked by the base selector. All rows, the base selector and the localization adapter included, come from localization passes that split a recording's partial tail block differently from the deployed ones, so they compare only with each other. TraceAV = accuracy over all its questions, hallucination sub-tasks included. \textbf{Bold} = best, \underline{underline} = second best in each column.}
\label{tab:retrieverkey}
\small
\setlength{\tabcolsep}{3.5pt}
\begin{tabular}{@{}lccccccc@{}}
\toprule
 & \multicolumn{4}{c}{Accuracy (\%)} & \multicolumn{3}{c}{Final-window coverage (\%)} \\
\cmidrule(lr){2-5}\cmidrule(lr){6-8}
Ranking score & TraceAV & LVOmni & VideoOdyssey & MMOU & TraceAV & VideoOdyssey & MMOU \\
\midrule
Base selector & 58.3 & 39.0 & 41.6 & 56.2 & 92.0 & 57.3 & 82.7 \\
Trial-answer confidence & 58.8 & 39.0 & 41.5 & 56.3 & 90.2 & 48.3 & 82.7 \\
SigLIP, frames & 58.8 & \underline{43.0} & 42.0 & 57.7 & \textbf{96.2} & \underline{65.8} & \underline{88.9} \\
bge-m3, transcript & \textbf{60.8} & 42.2 & \underline{43.6} & \underline{58.1} & \underline{96.0} & 58.9 & 88.3 \\
Localization adapter & \underline{59.2} & \textbf{43.3} & \textbf{45.6} & \textbf{61.1} & \textbf{96.2} & \textbf{71.7} & \textbf{94.8} \\
\bottomrule
\end{tabular}
\end{table}

\paragraph{An off-the-shelf retriever as the ranking score.} A third selector fits the same comparison, with its own base-selector and localization-adapter references (Table~\ref{tab:retrieverkey}). Keep the enumeration, the three-blocks-by-three-windows rule and the answerer, and rank windows by a retriever's similarity to the question: SigLIP image--text cosine over the window's frames, or bge-m3 text cosine against the window's speech transcript, the stem being the query in both (SigLIP truncates it to $64$ tokens).

Pooled over four benchmarks, both retrievers answer more accurately than the base selector, and the trained adapter is significantly ahead of the frame retriever and of the trial-answer score (\S\ref{app:ranking}). Its lead sits where localization training buys the most evidence coverage: on VideoOdyssey and MMOU it covers more annotated evidence than both retrievers and answers more accurately, significantly against both on MMOU and against the frame retriever alone on VideoOdyssey.

\paragraph{What the block count buys.} Evidence coverage rises with every block added, but accuracy does not follow it. With two blocks, \method answers significantly less accurately on CG-Bench and significantly more accurately on TraceAV. The block count is one setting for every benchmark, capped by what one answer pass holds. What still moves the answer at the deployed count is which blocks the ranking keeps: on VideoOdyssey, equally spaced blocks cost coverage and accuracy significantly (\S\ref{app:ranking}).

\subsection{The transcript channel: what it retains and what it answers}
\label{app:textregime}

\begin{table}[t]
\centering
\caption{\textbf{What each channel's selected windows retain on TraceAV.} \emph{Content-free uniform} spaces blocks and windows evenly and runs no pass. \emph{Media decoded} = what a scoring pass decodes at query time; \emph{tokens / block} = its measured length. \emph{Recall} = the share of annotated evidence spans the selected windows deliver, over all questions and over those whose recording tiles into more blocks than the retained-block count (\emph{ranking decides}). \textbf{Bold} = best, \underline{underline} = second best in each recall column.}
\label{tab:audiofirst}
\small
\begin{tabular}{@{}llrrr@{}}
\toprule
& & & \multicolumn{2}{c}{Recall} \\
\cmidrule(lr){4-5}
& & & All & Ranking \\
Channel & Media decoded & Tokens / block & questions & decides \\
\midrule
Content-free uniform & none & --- & 69.6 & 61.2 \\
Listen-only channel & audio & 6,855 & 71.7 & 67.0 \\
Media channel & frames + audio & 8,546 & \underline{74.1} & \underline{69.6} \\
Transcript channel & none & 1,579 & \textbf{74.3} & \textbf{70.0} \\
\bottomrule
\end{tabular}
\end{table}

\paragraph{What the selected windows retain.} Table~\ref{tab:audiofirst} reads TraceAV, whose questions are posed about audio and video together. Where the ranking decides (on recordings tiling into more blocks than the retained-block count, so block ranking has a choice), the transcript channel does not fall behind the media channel in mean per-question recall by more than a $2.5$\,pp non-inferiority margin. Neither the content-free uniform selection (equally spaced blocks and windows, no pass run) nor the listen-only channel (the localization pass fed audio alone) meets that margin. Within TraceAV the two channels separate along how much of the evidence is spoken. Each question's recall difference, transcript minus media, regressed on the speech share of its annotated evidence span rises by $+11.68$pp $[+5.80,+17.41]$ from evidence with no speech to evidence entirely spoken.

The same comparison on CG-Bench mini, a video benchmark whose human-labeled clue intervals mark visual evidence, runs the other way. Over all its questions the transcript channel retains significantly less of those intervals than the media channel, $-8.40$pp $[-10.54,-6.31]$, and the deployed answer pass reading its windows answers significantly less accurately, $-1.80$pp $[-2.97,-0.64]$.

\paragraph{The seconds of Table~\ref{tab:zoomend}.} The per-question seconds are a sampled replay on one idle H100, one question at a time, all media decoded fresh, at the deployed answer pass's eight-worker decode concurrency; the transcript is that of Appendix~\ref{app:details}, rendered per window under its character cap.

\subsection{The transcript outline at the answer pass}
\label{app:outline}

\begin{table}[t]
\centering
\caption{\textbf{The transcript outline at the answer pass.} Accuracy (\%). Every row answers the same retrieved windows, grouped by system. \emph{Question-ranked} (deployed) fills the outline question-first; \emph{Uniform} thins every minute equally; \emph{None} removes the outline. \textbf{Bold} = best, \underline{underline} = second best within each group. TraceAV = accuracy over all its questions, hallucination sub-tasks included.}
\label{tab:outline}
\small
\begin{tabular}{@{}llcccc@{}}
\toprule
System & Transcript outline strategy & TraceAV & LVOmni & VideoOdyssey & MMOU \\
\midrule
\multirow{3}{*}{\begin{tabular}[c]{@{}l@{}}\method without\\answer training\end{tabular}} & Question-ranked & \textbf{60.8} & \textbf{43.0} & \textbf{46.7} & \textbf{61.4} \\
 & Uniform & \underline{59.8} & \underline{42.8} & \underline{45.0} & 61.0 \\
 & None & 58.2 & 42.7 & 43.6 & \underline{61.1} \\
\midrule
\multirow{3}{*}{\begin{tabular}[c]{@{}l@{}}\method\end{tabular}} & Question-ranked & \textbf{63.6} & 45.8 & \textbf{53.7} & \textbf{66.3} \\
 & Uniform & \underline{63.2} & \underline{46.4} & 50.7 & \underline{66.2} \\
 & None & 60.4 & \textbf{47.5} & \underline{51.2} & 65.2 \\
\bottomrule
\end{tabular}
\end{table}

\begin{table}[t]
\centering
\caption{\textbf{Where the outline's gain lands.} Accuracy (\%) of \method, with the transcript outline present or removed. MMOU questions are grouped by how much of their annotated evidence interval the retained windows overlap; TraceAV questions by their annotated class.}
\label{tab:outlinegap}
\small
\begin{tabular}{@{}lcc@{}}
\toprule
Group & No outline & + transcript outline \\
\midrule
\multicolumn{3}{@{}l}{\emph{MMOU: overlap with the annotated evidence interval}} \\
Evidence uncovered & 48.0 & 55.9 \\
Partially covered & 63.8 & 65.5 \\
Covered (at least half) & 66.6 & 67.2 \\
\midrule
\multicolumn{3}{@{}l}{\emph{TraceAV: annotated question class}} \\
Spatiotemporal localization & 30.8 & 50.7 \\
All other classes & 63.8 & 65.1 \\
\bottomrule
\end{tabular}
\end{table}

\paragraph{Where the outline gains.} Table~\ref{tab:outlinegap} groups the questions by whether retrieval already delivered the evidence. On MMOU each question's annotated evidence interval, where the annotation gives a usable one, is intersected with the windows the deployed selection retained; the annotation is used only to group the answer pass and never reaches a prompt. The gain on the uncovered group is significant, from a starting accuracy far below the benchmark's overall accuracy, while the covered majority barely moves. On TraceAV the split is taken along its question classes instead, and the gain concentrates in the spatiotemporal-localization class.

\paragraph{Filling the outline question-first.} The deployed rule hands the outline's tokens to the minutes that score highest against the question; the \emph{Uniform} row of Table~\ref{tab:outline} instead thins every minute by the same fraction. A question that matches nothing, or an outline the cap never truncates, renders the same prompt under both strategies, so a benchmark-level row is diluted by however often the outline already fits. Read on the truncated questions alone, the \emph{Uniform} rule costs $3.57$ points on VideoOdyssey under \method and $1.81$ points on MMOU under \method without answer training, both significant. They are also the cells that stay significant at the benchmark level.

\subsection{The window grid}
\label{app:windows}

\begin{figure}[htbp]
\centering
\includegraphics[width=\textwidth]{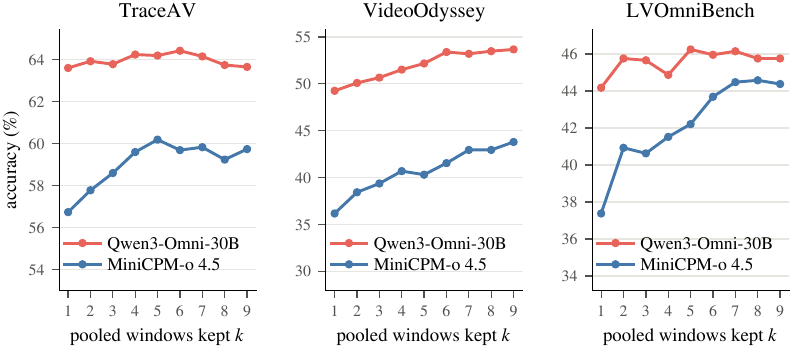}
\caption{\textbf{Accuracy against the answer-pass window count.} The pooled windows are truncated to the top-$k$ by window score. The deployed configuration keeps the whole pool, each curve's last point.}
\label{fig:keepcap}
\end{figure}

\paragraph{Where the deployed pool sits on the sweep.} A sensitivity sweep re-answers the deployed selection with the pooled window set truncated to its top-$k$ windows by window score, $k{=}1..9$, everything else held (Figure~\ref{fig:keepcap}). The ranking is global over the retained blocks, so the axis is the answer pass's total window count, with the per-block shortlist $W$ held at its deployed value throughout. It runs on both backbones on TraceAV, VideoOdyssey and LVOmniBench. On Qwen3-Omni-30B the TraceAV and LVOmniBench curves barely move from the first window on, and the VideoOdyssey curve rises until $k{\approx}6$ and barely moves after, every truncation to four or fewer windows reading significantly lower. On the smaller backbone the TraceAV curve reads significantly lower only at one and two windows; on VideoOdyssey its curve instead does not saturate, and every truncation of the pool reads lower, significantly so through the mid range. Its LVOmniBench curve rises until $k{\approx}6$, every truncation to four or fewer windows reading significantly lower.

\paragraph{Where the grid comes from.} The block length is set by the longest training clip: the localization adapter trains on clips that fit inside a single block (Appendix~\ref{app:repro}), and the longest clip in the LongVALE-derived corpus runs $597.9$\,s. At any larger $\Delta$ no training clip fills a block. With $\Delta$ fixed by the supervision, three axes stay free: the window width $\delta$, which sets $K{=}\lceil\Delta/\delta\rceil$ ($8$ at $\delta{=}75$\,s); the per-block shortlist $W$; and the retained-block count $B$.

\paragraph{The deployed grid survives a two-benchmark ablation.} We ablate the two axes Figure~\ref{fig:selection} leaves, $W$ and the window width, on CG-Bench mini and VideoOdyssey. \emph{Window count:} re-pooling the identical stored localization passes at $W{=}2$ barely moves accuracy on either benchmark, even though the dropped third window measurably costs clue-interval coverage on CG-Bench. \emph{Window width:} doubling the window to $150$\,s (at the same $W{=}2$, since three $150$\,s windows per retained block would overflow the answer-pass position cap) barely moves VideoOdyssey and drops significantly on CG-Bench. On CG-Bench both widths use a localization adapter retrained with single-level candidates at that width, so width is the only difference there; on VideoOdyssey the $75$\,s variant keeps the deployed adapter. The human clue intervals are consistent with dilution: the wider windows overlap the human clue intervals \emph{more} at the matched window count, yet answer less accurately.

\paragraph{Both counts are capped from above by one answer pass.} Adding a fourth block overflows one answer pass: a $4$-block configuration pools up to $12$ windows of $75$\,s, whose combined length reaches $36.9$--$45.3$k tokens on VideoOdyssey's overflowing questions against the deployed position cap of $36{,}864$. The add-a-block variants of Figure~\ref{fig:selection} are answered with the cap widened to $49{,}152$ for exactly the overflowing window sets. $W{=}4$ pools as many as twelve windows and meets the same cap.
\subsection{Where the windows sit: the random-placement control}
\label{app:randomwindow}

A control condition bounds the selection stage from below. For each question we take the retained windows, keep their number and each of their widths, and redraw their positions uniformly at random inside the recording (non-overlapping); the stored localization passes, the block ranking, the answering adapter and the per-window context stay the deployed configuration's own, so the window positions are the only difference. \method leads random placement by $1.41$ points on TraceAV (video-clustered 95\% CI $[+0.27,+2.55]$) and $4.99$ on VideoOdyssey $[+2.12,+7.83]$.

\subsection{The block-ranking score}
\label{app:ranking}

Where ranking has room to act, it is measurably better than equal spacing. Block ranking only has freedom on clips long enough to tile more than three blocks, so the benchmarks separate structurally: replacing the block ranking by equally spaced blocks changes the picked blocks on $95.5\%$ of VideoOdyssey's questions but on at most $51\%$ elsewhere. A paired comparison on VideoOdyssey (equally spaced blocks over the same stored localization passes, identical within-block windows, answering adapter and answer scoring) costs the equally spaced variant significantly (Figure~\ref{fig:selection}).

The benchmark's published per-question evidence spans say what that gain is made of. The equally spaced variant answers essentially as many window seconds per question as the deployed selection and still covers substantially less of the published evidence, a paired coverage gain of $23.07$ points for ranking $-$ equal spacing (video-clustered 95\% CI $[+18.48,+27.53]$).

The contrast sharpens exactly where placement should matter most: over the questions whose annotated evidence covers under a tenth of the clip the coverage gap widens to $29.49$ points, while the questions whose evidence sprawls across the clip, where almost any placement lands on it, dilute it.
\paragraph{What the ranking score is made of.} Because $\sigma$ is strictly increasing, $g_m=\max_k\sigma(\ell_{m,k})=\sigma(\max_k\ell_{m,k})$, so blocks are ranked exactly by their maximum logit, which splits exactly into the pass's \emph{average} logit $\bar{\ell}_m=\frac{1}{K_m}\sum_k\ell_{m,k}$, shared across the pass's windows, and the best window's \emph{margin} over it: $\max_k\ell_{m,k}=\bar{\ell}_m+(\max_k\ell_{m,k}-\bar{\ell}_m)$. A within-pass softmax deletes the average exactly, and subtracting it deletes it while keeping the $\max$; replacing the $\max$ by a mean keeps the average and discards the margin. The objective behind the logits matters in the same place: a selector trained with overlap-fraction BCE in place of Eq.~\ref{eq:localization_loss} ranks blocks worse across passes. Against a twin trained on the same corpus with Eq.~\ref{eq:localization_loss}, both on a single-level grid of $75$\,s windows, its highest-ranked block contains VideoOdyssey's annotated evidence $15.09$ points less often ($[-21.16,-9.66]$), and its top three blocks $13.87$ points less often. Answered by the deployed answer pass, the two selectors' accuracies lie within a point of each other: overlap-fraction BCE $-$ cross-entropy is $-0.36$ on TraceAV ($[-1.36,+0.64]$) and $+0.66$ on VideoOdyssey ($[-1.50,+2.94]$).

\begin{table}[t]
\centering
\caption{\textbf{The block-ranking score, ablated by component.} Each row re-ranks the same localization passes on VideoOdyssey. The deployed score is a monotone map of the block's average logit plus its best window's margin above it; \emph{Deletes} = the component a variant removes, a dash neither. A \emph{partial block} is a clip's tail block, with fewer candidate windows; \emph{partial block retained} = the share of questions retaining it. \textbf{Bold} = best, \underline{underline} = second best in each coverage column.}
\label{tab:rankingkey}
\small
\setlength{\tabcolsep}{5pt}
\begin{tabular}{@{}llccc@{}}
\toprule
& & \multicolumn{2}{c}{Evidence coverage (\%)} & Partial block \\
\cmidrule(lr){3-4}
Ranking score & Deletes & all questions & no partial block & retained (\%) \\
\midrule
$g_m=\max_k\sigma(\ell_{m,k})$ (deployed) & --- & \underline{70.2} & \textbf{78.7} & 67.6 \\
$\max_k\mathrm{softmax}(\ell)_k$ & average, renormalized & 63.8 & 73.8 & 62.5 \\
$\max_k\ell_{m,k}-\bar{\ell}_m$ & average, subtracted & \textbf{72.9} & \underline{78.0} & 11.8 \\
$\bar{\ell}_m$ & margin & 33.8 & 39.0 & 82.1 \\
$\tfrac{1}{K_m}\sum_k\sigma(\ell_{m,k})$ & margin; mean of $\sigma$ & 23.8 & 26.8 & 77.8 \\
\bottomrule
\end{tabular}
\end{table}

\paragraph{Deleting the average or the margin.} Which ranking retains the \emph{better} blocks is scored directly: each is read on evidence coverage over the identical stored localization passes (Table~\ref{tab:rankingkey}). Within a block the window shortlist is taken under a monotone map of the same logits and is therefore invariant to the average, so re-ranking changes which blocks are retained and nothing else. Deleting the average moves coverage by single digits, in a direction set by how it is removed. Deleting the margin, by a mean in place of the $\max$, costs tens of coverage points, and costs coverage significantly even on the questions with no partial tail block; the literal substitution $\frac{1}{K_m}\sum_k\sigma(\ell_{m,k})$ for $g_m$ sits lowest, since with every letter's sigmoid near one it ranks a block by its \emph{weakest} windows. What the ranking runs on is the margin. The deployed score $\max_k\sigma(\ell_{m,k})$ is its plainest form, a monotone rescaling of the strongest candidate-window logit with no normalization step of its own.

\paragraph{Trial-answer scores.} A block can also be scored by answering the question on it and reading the chosen option's probability~\citep{shen2025zoom}. Read on the block as the localization pass sees it, that score costs what one localization pass costs; in our pipeline it is a second pass on every block, since the window logits still come from the localization pass, taking the search from $M{+}1$ passes to $2M{+}1$. It is renormalized within its pass, so it discards the shared average as the softmax variant does. What survives is mostly about the question. A block is only ever ranked against the other blocks of its own question, so a ranking uses only the within-question part of a score's variation. Over the stored trial answers of the $30$\,s-window scan (\S\ref{app:selection}) $71\%$ of that probability's variance sits between questions, against $29\%$ for the maximum window logit behind $g_m$. The two scores rank differently: on questions whose recordings are long enough to rank, they agree on the best block $30\%$ of the time.

Put in the localization pass's place on the protocol of Table~\ref{tab:retrieverkey}, with windows inside a block ranked by the base selector, the trial-answer score covers less annotated evidence than the base selector on VideoOdyssey and TraceAV and stays within a fraction of a point of its accuracy on every benchmark. The trained adapter answers more accurately than it on LVOmniBench, VideoOdyssey and MMOU.

\subsection{Selection hits convert to correct answers: CG-Bench and the base-selector control}
\label{app:cgbench}

CG-Bench mini (3,000 questions over 1,118 videos of $9$--$105$ minutes) ships \emph{per-question human-labeled clue intervals} at second scale: an external annotation of the quantity the selector should find, on clips nearly all of which fit one whole-clip pass. Against the whole clip with the identical answer LoRA, \method wins the paired comparison significantly (57.50 against 49.53).

\paragraph{Official-protocol robustness.} CG-Bench's official evaluation differs from ours in four ways: it feeds frames and subtitle text with no audio, where ours keep the native audio; its prompt carries the subtitles (with per-cue timestamps) and the sampled-frame timestamps, plus a JSON answer format with a forced-guess instruction; when no JSON parses, its letter extraction scores the gold letter anywhere in the response as correct; and both official scripts drop unanswered rows. The lenient extraction changes none of our answers, and dropping unanswered rows moves either system by less than half a point. Rerunning \method and the whole clip under the official prompt (for \method, its subtitle block cut to the retained windows takes the place of its transcript outline, without the per-frame timestamp list) and scoring with the official scripts gives \method $56.57$ and the whole clip $50.42$ on the questions each answers; the lead holds with unanswered questions counted wrong.

\begin{figure}[htbp]
\centering
\includegraphics[width=0.71\textwidth]{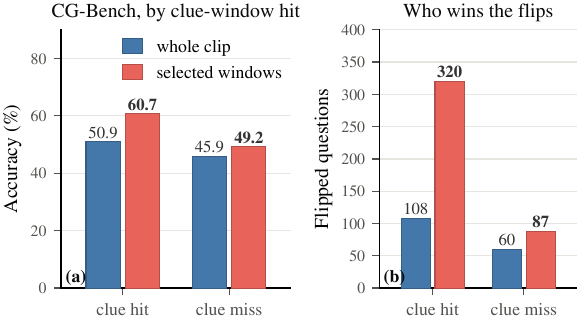}
\caption{\textbf{Clue-grounded conversion on CG-Bench.} \method against the whole clip; questions split by whether the selected windows overlap the human-labeled clue interval (\emph{hit}) or not (\emph{miss}). \emph{(a)} Accuracy within each stratum. \emph{(b)} Discordant flips, where exactly one of the two is correct.}
\label{fig:cgbench_clue}
\end{figure}

The human clue intervals test, descriptively, where that lead comes from. The selected windows overlap the labeled clue interval (a \emph{hit}) on $72.3\%$ of questions, and where they do the same answer pass is markedly more often right. Stratifying the paired comparison by this external hit criterion (Figure~\ref{fig:cgbench_clue}) puts most of the lead over the whole clip inside the hit stratum, and the paired flips break the same way: lopsidedly for \method where the windows hit, far less so where they miss.

\paragraph{The same ordering in the leaderboard's own aggregate (CRR).} CG-Bench's aggregate counterpart to the stratification above is \emph{CRR} $= \min(\text{long-acc},\,\text{clue-acc})/\text{clue-acc}$, with long-acc the accuracy on the whole video; it requires a \emph{clue-acc} condition: the same answer LoRA and transcript outline, with the media restricted to the human clue intervals, under the official clue prompt at the official frame count of $32$, with at least two frames per clue interval. It puts clue-acc at $64.47$. With long-acc taken from the paired comparison above, over all 3,000 questions, \method realizes $\mathrm{CRR}{=}0.892$ and the whole clip $0.768$, the same ordering the stratified reading gives.

\begin{figure}[htbp]
\centering
\includegraphics[width=\textwidth]{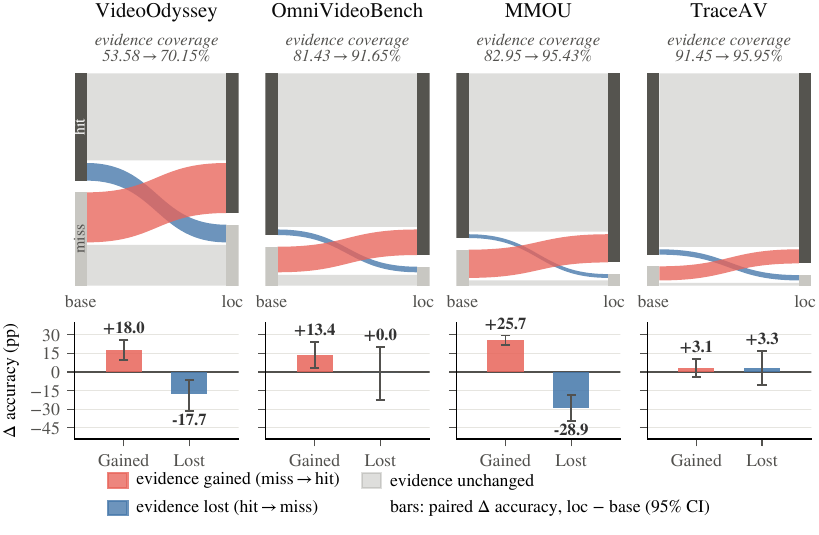}
\caption{\textbf{Evidence-hit conversion inside the base-selector control.} \emph{Top}: each question flows from its final-window evidence status under the base selector (left) to its status under the localization LoRA (right); ribbon thickness is the share of questions on that transition, grey the unchanged. \emph{Bottom}: the paired accuracy change, localization minus base, on the questions of each coloured transition, with its video-clustered interval. OmniVideoBench is read on its timestamped subset, MMOU on the questions with a stored selection under both selectors; LVOmniBench ships no timestamps.}
\label{fig:loc_conversion}
\end{figure}

\paragraph{The internal companion: the same stratification inside the base-selector control.} Four of the five AV benchmarks of Figure~\ref{fig:locablation} ship evidence timestamps, so the same question can be asked \emph{inside} its selector ablation, where only the selector adapter varies. Classify every question by whether a final window covers its annotated evidence under the base selector and under the localization LoRA (Figure~\ref{fig:loc_conversion}). On MMOU, VideoOdyssey and OmniVideoBench the paired accuracy gain concentrates substantially on the questions where the trained selector \emph{gains} the hit, is smaller where the hit status does not change, and \emph{reverses} where the trained selector loses the hit, on VideoOdyssey and MMOU, most sharply on MMOU. Across benchmarks, the final-window coverage left to gain spans $8.55$ to $46.42$ points, and the paired accuracy gain per coverage point gained spans $0.18$ to $0.44$.

\subsection{Audio at both stages of the pipeline}
\label{app:audio}

\begin{table}[htbp]
\centering
\caption{\textbf{The audio ablation on TraceAV-Bench, at the two stages it can be read.} (a) silences the localization pass alone and scores what the selected windows retain, with no answer pass; (b) keeps the audio-on selection's windows, silences their answer pass (audio and transcript outline), and scores the answers.}
\begin{subtable}[t]{0.5\textwidth}
\centering
\caption{\textbf{What the localization pass's audio buys the selected windows.} An evidence event counts as retained when a selected window overlaps its annotated span; events are grouped by whether the annotation says hearing carries them, alone or together with seeing, or seeing alone; every question is read.}
\label{tab:evhit}
\small
\setlength{\tabcolsep}{3pt}
\begin{tabular}{lcc}
\toprule
Evidence spans retained & Audio on & Muted \\
\midrule
Carried by hearing & 75.1 & 61.5 \\
Carried by seeing & 71.2 & 65.6 \\
\bottomrule
\end{tabular}
\end{subtable}

\hfill
\begin{subtable}[t]{0.48\textwidth}
\centering
\caption{\textbf{What the selected windows hear.} The same windows and the same answer LoRA in both conditions, cut by the benchmark's own question-modality class, its code in parentheses.}
\label{tab:audioablation}
\small
\setlength{\tabcolsep}{2.5pt}
\begin{tabular}{lcc}
\toprule
Question class & Audio on & Muted \\
\midrule
Hearing-required (\texttt{a\_}) & 71.9 & 66.5 \\
Cross-modal (needs both) (\texttt{av\_}) & 59.6 & 47.7 \\
Vision-only (\texttt{v\_}) & 44.0 & 43.2 \\
Fabricated premise (\texttt{mh\_}) & 76.6 & 74.2 \\
\bottomrule
\end{tabular}
\end{subtable}

\end{table}

\paragraph{Audio inside the retrieval stage.} On TraceAV, silencing the \emph{localization pass} alone returns visibly different windows: the median window-set Jaccard against audio-on is $0.43$ (IQR $0.29$--$0.64$).

The changed selection can be scored directly, without an answer pass: every TraceAV question ships an evidence trajectory whose events carry both a time range and their own modality label. Scoring each selection on whether it still contains those spans separates the two conditions cleanly (Table~\ref{tab:evhit}). Muting the localization pass costs both, but it costs the hearing-borne evidence substantially more, each loss significant on its own video-clustered interval, and the audio-specific excess is significant in its own right. What the localization pass's audio buys is mostly the retrieval of the spans the annotation marks as heard, alone or together with seeing.

\paragraph{The labels those readings rest on.} TraceAV is the benchmark whose questions ship those labels (answerable by hearing alone, by seeing alone, by both, or a fabricated premise the model should refuse), and its paper validates them on this backbone: its visual-only ablation reruns Qwen3-Omni-30B-A3B with audio removed at the official recipe, and the audio-centric sub-tasks fall sharply while the visual-centric ones barely move~\citep{feng2026traceav}. We read \method's answer-pass silencing against that validation, over one fixed set of windows (Table~\ref{tab:audioablation}): the cost is largest on the cross-modal and hearing-required classes and smallest on the vision-only class.

\paragraph{Muting both passes, per length bucket.} The muted condition, with no audio and no transcript outline, is drawn against its audio-fed twin bucket by bucket in the LVBench and Video-MME panels of Figure~\ref{fig:length_appendix}. Audio, with the outline built from its transcript, is load-bearing there, and on Video-MME its value grows with duration: muting costs several times more on the 30--60-minute bucket than on the sub-5-minute one.

\FloatBarrier
\section{The Answering Stage and the Cross-Backbone Transfer}
\label{app:answer}

\paragraph{The no-media condition.} The no-media condition of Figure~\ref{fig:locablation} re-runs the answer prompts of the two selector variants without media: stem and options only, greedy, the same letter parser as every condition, the localization adapter answering as in the selector variants, and the same one-line answer instruction. The localization adapter is more accurate than this condition on every benchmark.

\begin{figure}[t]
\centering
\includegraphics[width=\textwidth]{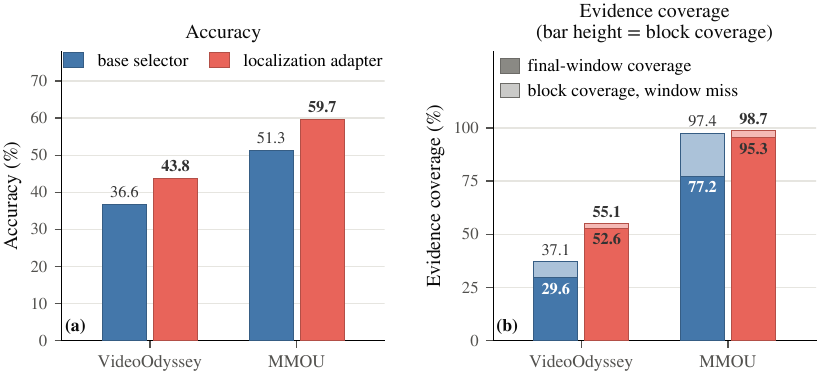}
\caption{\textbf{The retrieval stage ablated on MiniCPM-o 4.5.} Only the localization adapter differs. \emph{(a)} Accuracy. \emph{(b)} Evidence coverage; full bar = block coverage, saturated part = final-window coverage.}
\label{fig:locablationport}
\end{figure}
\paragraph{Localization on the smaller backbone.} Figure~\ref{fig:locablationport} runs the selector ablation of Figure~\ref{fig:locablation} on MiniCPM-o 4.5, on VideoOdyssey and MMOU: the same block grid, the same candidate-window grid and block ranking, the smaller backbone's own answer adapter in both variants, and only the localization pass's adapter swapped. Training the selector raises final-window coverage substantially on both benchmarks; block coverage moves substantially only on VideoOdyssey, where the hours-long recordings leave the block ranking room to move, and stays near ceiling on MMOU, most of whose recordings fit one block. Paired accuracy rises significantly on both, on MMOU by more than the same swap buys the larger backbone, from a lower untrained starting point.

\begin{figure}[htbp]
\centering
\includegraphics[width=0.55\textwidth]{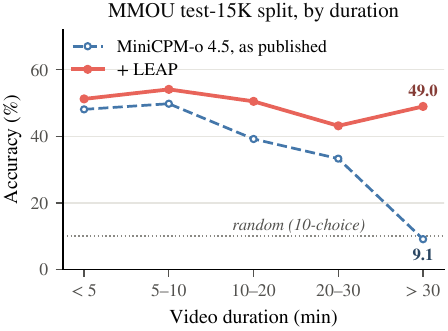}
\caption{\textbf{MMOU by video duration: MiniCPM-o 4.5 on the test-15K split.} Scored per duration bucket by the benchmark's own evaluator. Dashed = the duration breakdown the benchmark paper reports for its own MiniCPM-o 4.5 run; solid = the same backbone carrying \method.}
\label{fig:mmou_duration}
\end{figure}

\paragraph{Against the published rows for the same backbone.} Three of the smaller backbone's benchmarks publish their own MiniCPM-o 4.5 row (Table~\ref{tab:minicpm-official}). MMOU withholds the gold labels of its test-15K split and scores submissions with its own evaluator, so every row of that column is the benchmark's own score. Outside this column, Figure~\ref{fig:mmou_duration} and the MMOU axis of Figure~\ref{fig:benchradar}b, every MMOU number in this paper is on the benchmark's $5{,}000$-question test-mini split, whose gold labels are public; the two splits share no question. We submitted the deployed MiniCPM-o 4.5 configuration on all $15{,}000$ questions, answering each once in the original option order; questions the pipeline leaves unanswered are kept in the submission, so the denominator is the full split. The published row's protocol additionally votes over five option-order shuffles and selects among prompt variants, neither of which we use. That evaluator also returns a duration breakdown (Figure~\ref{fig:mmou_duration}): \method is ahead in every bucket, and its accuracy on the longest videos stays near its overall one.

\paragraph{The agent of its size class.} Our OmniAgent-RL-7B row drives the released checkpoint with its own evaluator at its defaults: a duration-adaptive cap of at most thirty-two steps, sampled decoding, and our extractor for scoring. TraceAV's appendix drives the same checkpoint with the benchmark's harness and scorer under a thirty-two-step cap (Table~\ref{tab:minicpm-official}). \method leads both readings.

\begin{table}[t]
\centering
\caption{\textbf{The smaller backbone against its published rows and two systems of its size class.} Accuracy (\%). \textsuperscript{*}~= published number \citep{feng2026traceav,tao2026lvomnibench,goel2026mmou}. \textsuperscript{a}~= a re-watching system on Qwen3-VL-8B with a Whisper encoder, as its paper reports it \citep{li2026salmonnr3}. \textsuperscript{b}~= a native omni agent on Qwen2.5-Omni-7B, run by us under its own evaluator and step limit \citep{xing2026omniagent}. --- = no number. Columns follow each publication's protocol: TraceAV's two tables as subtask macro-averages, LVOmni micro, MMOU's test-15K split under the benchmark's own evaluator. \textbf{Bold} = best, \underline{underline} = second best in each column.}
\label{tab:minicpm-official}
\small
\begin{tabular}{lcccc}
\toprule
System & TraceAV & TraceAV & LVOmni & MMOU \\
 & General & Hallucination &  & test-15K \\
\midrule
video-SALMONN-R$^3$ (8B)\textsuperscript{a} & --- & --- & \underline{42.9} & --- \\
OmniAgent-RL-7B\textsuperscript{b} & 47.6 & 42.0 & 39.4 & 31.6 \\
OmniAgent-RL-7B, as published\textsuperscript{*} & \underline{55.0} & 46.1 & --- & --- \\
\midrule
MiniCPM-o 4.5, as published\textsuperscript{*} & 44.8 & \underline{66.5} & 34.8 & \underline{46.8} \\
~~+ \method & \textbf{57.8} & \textbf{69.6} & \textbf{44.4} & \textbf{51.0} \\
\bottomrule
\end{tabular}
\end{table}

\FloatBarrier
\section{Where the Advantage Comes From: Duration, Certificate Length, and Evidence Position}
\label{app:regime}

\subsection{Accuracy across video length}
\label{app:length}

\begin{figure}[t]
\centering
\includegraphics[width=\textwidth]{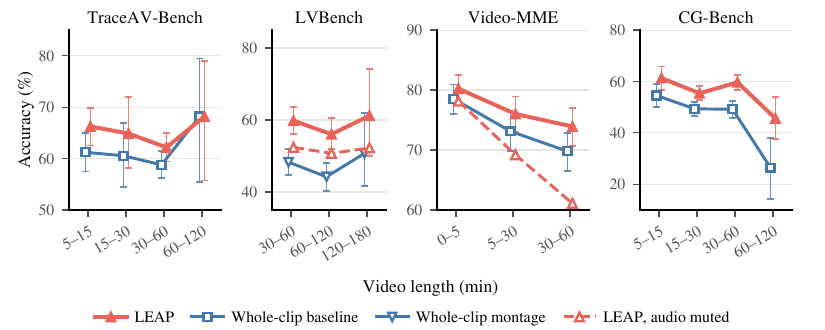}
\caption{\textbf{Accuracy vs.\ video length.} Video-MME pools its two middle buckets. Muted = audio off at both passes, no transcript outline. Bars = video-clustered 95\% interval, none on the muted line.}
\label{fig:length_appendix}
\end{figure}

The three video benchmarks' official protocols are visual, and LVBench's own distribution ships no audio, whereas our runs feed native audio and video, LVBench's from local copies of its source videos (video only on the questions whose copy carries no audio stream). Figure~\ref{fig:benchradar} reads LVBench and Video-MME with audio removed throughout; without audio every clip fits one whole-clip pass, so the baseline of Figure~\ref{fig:benchradar}a there is the whole clip. Figure~\ref{fig:length_appendix} reads them with native audio, its dashed line being the muted condition. Every whole-clip and montage baseline on them carries the answer LoRA. In Figure~\ref{fig:length_appendix}, each panel's baseline follows \S\ref{sec:results}: the whole clip on Video-MME and CG-Bench, where $\ge$$99.3\%$ of questions fit one whole-clip pass, and the context-filled montage on LVBench, where questions exceed the whole-clip limit while the montage always fits.

On MMOU \method rises monotonically with duration. Up to thirty minutes \method shares part of that rise with the whole clip read over the \emph{same questions}; past thirty minutes \method holds its level and its lead over the whole clip widens (Figure~\ref{fig:length}).

On the three video benchmarks, the per-bucket detail behind \S\ref{sec:beyond}: on LVBench \method leads the context-filled montage in every bucket; on CG-Bench \method leads the whole clip in every bucket; on Video-MME with native audio \method leads the whole clip in every bucket, the more so the longer the recording. The TraceAV panel of Figure~\ref{fig:length_appendix} covers only questions inside the whole-clip limit.

\subsection{Inside VideoOdyssey: certificate length}
\label{app:vo_inside}

VideoOdyssey annotates every question with the length of the continuous span of the recording that certifies its answer, and it ships both a vision-only and an audio-visual track. Table~\ref{tab:vo_ccl} sorts both tracks by that length. Its rows are matched in frame count. Our variant re-answers the deployed selection with the answer LoRA at a per-window frame cap sized so that nine windows approach the native-sampling whole clip's frame count, at its pixel size, and without the transcript outline.

Both tracks are lifted most at the two ends of that axis and least in the interior, and the two ends admit two different readings, tested below. On the audio-visual track the sharpest gain sits where the certificate is shortest; one reading is evidence locality: the answer lives in a small span of a long recording, and a trained selector recovers it at the per-window context. The other is context pressure at the longest certificates: evidence that extensive is landed on by almost any placement, and what remains is the advantage of windows read at the per-window context over a baseline stretched thinnest across these recordings. The interior is where both readings predict the least.

\begin{table}[t]
\centering
\caption{\textbf{Accuracy by certificate length on both VideoOdyssey tracks.} \emph{Certificate length} = the benchmark's annotated length of the continuous span a viewer must watch to answer. $\dagger$ = the native-sampling whole clip on the same backbone, on the audio-visual track with the benchmark's official audio montage. The \method rows re-answer their selected windows at $14$ frames per window, matching the $\dagger$ frame count, and without the transcript outline. \textbf{Bold} = best per column and track.}
\label{tab:vo_ccl}
\small
\setlength{\tabcolsep}{5pt}
\begin{tabular}{lcccccc}
\toprule
System & \multicolumn{5}{c}{Certificate length (min)} & Overall \\
\cmidrule(lr){2-6}
 & [0, 0.5) & [0.5, 3) & [3, 15) & [15, 60) & [60, $\infty$) & \\
\midrule
\multicolumn{7}{l}{\emph{VideoOdyssey-V (vision-only)}} \\
\midrule
Qwen3-Omni-30B base$^\dagger$ & 38.6 & 44.5 & \textbf{44.8} & \textbf{40.4} & 35.8 & 41.2 \\
\method & \textbf{44.0} & \textbf{49.5} & 41.3 & \textbf{40.4} & \textbf{43.1} & \textbf{44.1} \\
\midrule
\multicolumn{7}{l}{\emph{VideoOdyssey-AV (audio-visual)}} \\
\midrule
Qwen3-Omni-30B base$^\dagger$ (+ audio montage) & 34.0 & 33.7 & 43.0 & 35.3 & 38.9 & 36.7 \\
\method & \textbf{54.5} & \textbf{48.4} & \textbf{54.0} & \textbf{38.3} & \textbf{55.6} & \textbf{50.5} \\
\bottomrule
\end{tabular}
\end{table}

\begin{table}[t]
\centering
\caption{\textbf{The two readings of the certificate-length axis, tested on VideoOdyssey-AV.} \emph{Certificate hit} = an answered window overlaps the annotated certificate span. \emph{Random placement} keeps the count of \method's selected windows and redraws their positions over the recording. \textbf{Bold} = best in each column within a panel.}
\label{tab:vo_ccl_mech}
\small
\begin{tabular}{lccccc}
\toprule
 & \multicolumn{5}{c}{Certificate length (min)} \\
\cmidrule(lr){2-6}
 & [0, 0.5) & [0.5, 3) & [3, 15) & [15, 60) & [60, $\infty$) \\
\midrule
\multicolumn{6}{l}{\emph{Certificate hit (\%)}} \\
\midrule
\method & \textbf{60.7} & \textbf{69.2} & \textbf{65.8} & 85.0 & 95.8 \\
Random placement & 9.4 & 20.5 & 43.0 & \textbf{88.7} & \textbf{100.0} \\
\midrule
\multicolumn{6}{l}{\emph{Accuracy (\%)}} \\
\midrule
\method & \textbf{55.7} & \textbf{49.6} & \textbf{53.3} & 37.6 & \textbf{61.1} \\
Random placement & 47.1 & 39.0 & 47.4 & \textbf{38.3} & 54.2 \\
Context-filled montage & 39.3 & 39.3 & 47.8 & \textbf{38.3} & 47.2 \\
\bottomrule
\end{tabular}
\end{table}

\paragraph{Testing the two readings.} Table~\ref{tab:vo_ccl_mech} tests both on the deployed selection, at the deployed $32$ frames per window; the \method row of Table~\ref{tab:vo_ccl} answers the same windows at $14$ frames each, which is why the two \method rows differ. It sets that selection against a random placement of as many windows, drawn as in Appendix~\ref{app:randomwindow} and answered without the transcript outline like every row here, and against the context-filled montage read by the same answer adapter. Evidence locality predicts that placement matters where the certificate is short. There \method's windows hit the certificate far more often than random windows, and answer significantly more accurately. Past fifteen minutes random placement hits the certificate about as often. Context pressure predicts a lead that remains once placement stops mattering. At the longest certificates nearly every placement hits, the recordings are the longest of the five levels, and \method still leads the context-filled montage significantly.

\begin{figure}[t]
\centering
\includegraphics[width=0.6\textwidth]{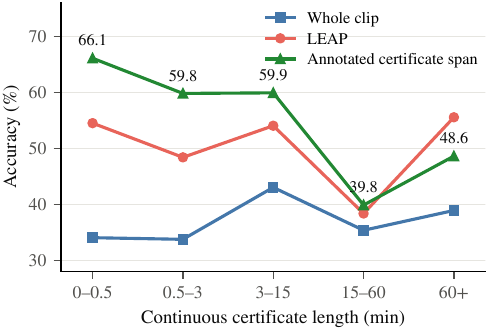}
\caption{\textbf{The annotated certificate span on VideoOdyssey-AV, by certificate length.} \emph{Whole clip} = the native-sampling whole clip. \method reads $14$ frames per selected window. A moment-annotated question is absent from the certificate condition, the only one labelled with values.}
\label{fig:vo_cw}
\end{figure}

\paragraph{What the annotated certificate is worth.} The benchmark also asks what happens when the answer pass reads that certifying span instead of the recording, and Figure~\ref{fig:vo_cw} runs it on this stack, sampled like the native-sampling whole clip ($2$ frames per second, at most $128$ frames). \method's selected windows trail it significantly at the two shortest bins, and the gap closes as the certificate widens. Past fifteen minutes the certifying span outlasts the eleven minutes of audio an official-style montage carries, so the span arrives with the audio subsampling the native-sampling whole clip already carries.

\subsection{Inside VideoOdyssey: the task types}
\label{app:vo_tasks}

The benchmark also labels every question with one or more of eighteen task types. Table~\ref{tab:vo_tasks} adds the \method row to its own per-task comparison, read against the same-stack baseline (\S\ref{app:samestack}).

\method's per-task accuracy is above that baseline's on every task type except Counting.

\begin{table}[t]
\centering
\caption{\textbf{Per-task accuracy on the audio-visual track of VideoOdyssey.} Columns are the benchmark's task types; a question annotated with several is scored under each. \emph{Count} counting, \emph{ObRec} object, \emph{AcRec} action, \emph{VAR} visual attribute, \emph{AER} acoustic event, \emph{AAR} acoustic attribute and \emph{OCR} character recognition, \emph{SFR} speech fact retrieval, \emph{Cap} captioning, \emph{CaRea} causal, \emph{EmRea} emotional, \emph{InRea} intentional and \emph{ObRea} object reasoning, \emph{SCR} speech content reasoning, \emph{SpRea} spatial reasoning, \emph{Order} temporal ordering, \emph{Sum} summarization, \emph{TeGro} temporal grounding. \textsuperscript{\ddag} = the benchmark's official input recipe re-run on our stack; rows above it are published values. \textbf{Bold} = best, \underline{underline} = second best in each column, the human row excluded.}
\label{tab:vo_tasks}
\setlength{\tabcolsep}{2pt}
\resizebox{\textwidth}{!}{%
\begin{tabular}{lccccccccccccccccccc}
\toprule
 & \multicolumn{9}{c}{\textbf{Perception}} & \multicolumn{7}{c}{\textbf{Cognition}} &  &  &  \\
\cmidrule(lr){2-10} \cmidrule(lr){11-17}
System & Count & ObRec & AcRec & VAR & AER & AAR & OCR & SFR & Cap & CaRea & EmRea & InRea & ObRea & SCR & SpRea & Order & Sum & TeGro & Overall \\
\midrule
\multicolumn{20}{l}{\emph{Human baseline}} \\
\midrule
Human & 75.0 & 85.0 & 85.4 & 81.0 & 78.7 & 71.1 & 81.1 & 87.9 & 80.6 & 81.0 & 74.4 & 79.3 & 73.7 & 75.0 & 82.6 & 71.0 & 70.8 & 93.2 & 80.7 \\
\midrule
\multicolumn{20}{l}{\emph{Proprietary omni-modal LLMs}} \\
\midrule
Gemini-2.5-Pro & 25.9 & 41.3 & 40.0 & 44.8 & \underline{38.4} & 44.6 & 45.8 & 50.3 & \underline{74.5} & 50.0 & 46.8 & \underline{48.1} & \textbf{42.9} & \underline{53.2} & 23.3 & 38.0 & 60.0 & 37.7 & 43.9 \\
Gemini-3-Flash & 30.9 & 44.4 & \underline{45.0} & 36.2 & 32.6 & 46.2 & 45.8 & 50.9 & 66.0 & \underline{56.9} & \underline{51.6} & \underline{48.1} & \textbf{42.9} & \underline{53.2} & 30.0 & \textbf{44.0} & 50.0 & 32.8 & 44.3 \\
Gemini-3.1-Pro & 28.4 & \textbf{52.4} & 40.0 & 43.1 & 37.2 & 41.5 & \underline{52.1} & \underline{59.5} & 61.7 & 53.5 & 46.8 & \textbf{65.4} & \underline{41.1} & 41.9 & 28.3 & 34.0 & \underline{64.0} & 39.3 & \underline{46.1} \\
Qwen3.5-Omni-Plus & \underline{37.0} & \underline{50.8} & 38.3 & \underline{46.6} & \underline{38.4} & \underline{52.3} & 33.3 & 43.4 & 72.3 & 50.0 & 48.4 & 40.4 & 35.7 & 45.2 & \underline{31.7} & 20.0 & 62.0 & 42.6 & 43.0 \\
\midrule
\multicolumn{20}{l}{\emph{Open-source omni-modal LLMs}} \\
\midrule
Qwen3-Omni-30B & 24.7 & 34.9 & 36.7 & 22.4 & 25.6 & 29.2 & 29.2 & 28.9 & 38.3 & 29.3 & 29.0 & 26.9 & 23.2 & 29.0 & 16.7 & 18.0 & 34.0 & 37.7 & 28.7 \\
Qwen3-Omni-30B (official recipe)\textsuperscript{\ddag} & \textbf{38.3} & 46.0 & 40.0 & 37.9 & 36.0 & 35.4 & 35.4 & 39.9 & 66.0 & 37.9 & 33.9 & 30.8 & 19.6 & 24.2 & 26.7 & 26.0 & 54.0 & \underline{44.3} & 36.9 \\
\method & 34.6 & 49.2 & \textbf{50.0} & \textbf{50.0} & \textbf{47.7} & \textbf{53.8} & \textbf{60.4} & \textbf{72.8} & \textbf{78.7} & \textbf{60.3} & \textbf{53.2} & \underline{48.1} & 39.3 & \textbf{61.3} & \textbf{36.7} & \underline{40.0} & \textbf{66.0} & \textbf{52.5} & \textbf{53.7} \\
\bottomrule
\end{tabular}}
\end{table}

\subsection{The lead over the montage across its audio coverage}
\label{app:montage}

\begin{figure}[tp]
\centering
\includegraphics[width=\textwidth]{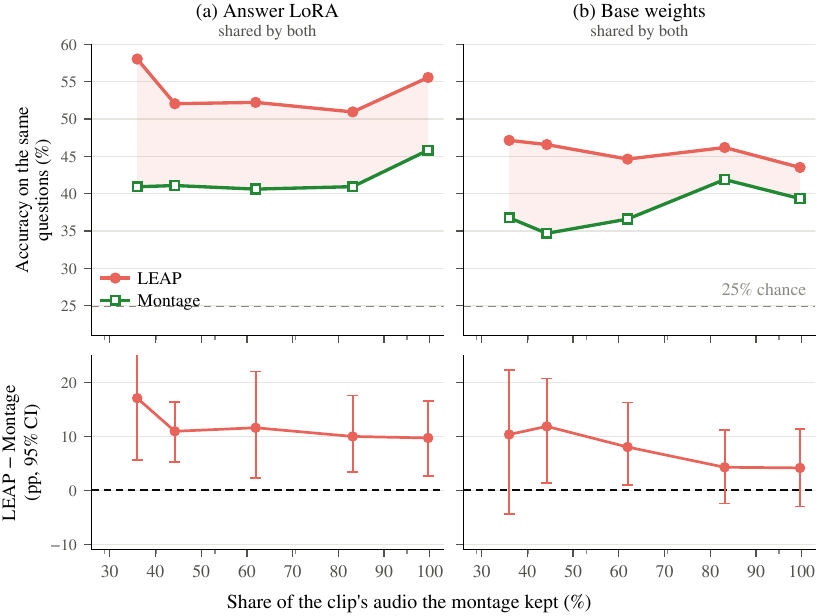}
\caption{\textbf{Accuracy by montage audio coverage, VideoOdyssey-AV.} Questions pool into coverage quantile bins, points at bin means, edges as minor ticks. Both systems answer with the answer LoRA in \emph{(a)} and with the base weights in \emph{(b)}. \emph{Bottom}: paired difference.}
\label{fig:montage_cov}
\end{figure}

The five bins of Figure~\ref{fig:montage_cov} are cut on the context-filled montage's own audio coverage, which falls with duration. With the same answer LoRA on both, \method leads the montage significantly in every bin, including the one where the montage keeps almost all of the audio. With the base weights on both, \method still leads in every bin, significantly in the second and third.

\subsection{Where the evidence sits in the clip}
\label{app:position}

\begin{figure}[tp]
\centering
\includegraphics[width=0.88\textwidth]{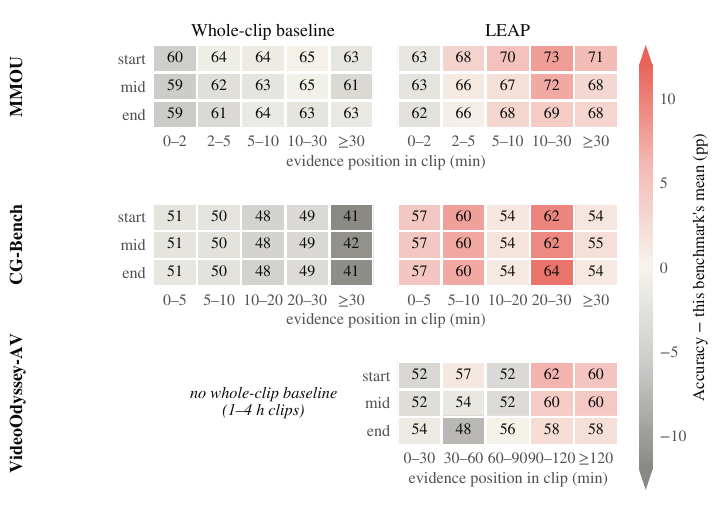}
\caption{\textbf{Accuracy by evidence position, over the questions carrying evidence timestamps.} The rows reduce a question's annotated evidence span to one instant: its start, midpoint or end. \method and the whole clip carry the answer LoRA. Cells print bucket accuracy.}
\label{fig:niah}
\end{figure}

Figure~\ref{fig:niah} splits \method and the whole clip by evidence position. On MMOU \method climbs across the axis and leads by more once the evidence sits past the tenth minute; on CG-Bench it leads in every bucket. On VideoOdyssey \method does not decay across evidence positions spanning hours.

\end{document}